\documentclass[journal]{IEEEtran}

\usepackage{cite}
\usepackage[pdftex]{graphicx}
\usepackage{amsmath}
\usepackage{amsthm}
\usepackage{amsfonts}
\usepackage{amssymb}
\usepackage{mathtools}
\usepackage{graphicx}
\usepackage{color}
\usepackage{array} 
\usepackage[caption=false,font=footnotesize]{subfig}
\usepackage{stfloats} 
\usepackage{url}
\usepackage{mathtools}
\usepackage{xcolor}

\usepackage{pgfplots}
\pgfplotsset{compat=newest}
\usetikzlibrary{plotmarks}
\usepgfplotslibrary{patchplots}
\usepackage{grffile}

\pgfplotsset{
	compat=newest,
	/pgfplots/myylabel absolute/.style={%
		/pgfplots/every axis y label/.style={at={(0,0.5)},xshift=#1,rotate=90},
		/pgfplots/every y tick scale label/.style={
			at={(0,1)},above right,inner sep=0pt,yshift=0.3em
		}
	}
}

\pgfplotsset{plot coordinates/math parser=false}
\pgfplotsset{every axis legend/.append style={
		at={(0.7,0.9)},
		anchor={north west},
		nodes={scale=0.5, transform shape}}}
\newlength\figureheight
\newlength\figurewidth

\graphicspath{{./}}

\theoremstyle{plain}

\theoremstyle{definition}

\usepackage{algorithm}
\usepackage{algpseudocode}
\algnewcommand\algorithmicAnd{\textbf{and} }
\algnewcommand\algorithmicOr{\textbf{or} }
\algrenewcommand{\algorithmiccomment}[1]{\hskip1em$\setminus\setminus$ #1}

\def\CC{{C\nolinebreak[4]\hspace{-.05em}\raisebox{.4ex}{\tiny\bf ++}}}

\let\vec\boldvec
\newcommand\at[2]{\left.#1\right|_{#2}} 
\DeclareMathOperator{\vect}{vec}

\newcommand{\joint}{\vec{q}} 
\newcommand{\state}{\vec{x}} 
\newcommand{\error}{\vec{e}} 
\newcommand{\traj}{\vec{r}} 

\newcommand{\linDist}{\vec{d}} 

\newcommand{\sysInput}{\vec{u}} 
\newcommand{\linInput}{\vec{\delta u}} 
\newcommand{\trjInput}{\sysInput_{\mathrm{IDM}}} 

\newcommand{\hitTime}{T} 

\newcommand{\hitFun}{\vec{\Psi}_{\mathrm{hit}}} 
\newcommand{\landFun}{\vec{\Psi}_{\mathrm{land}}} 
\newcommand{\netFun}{\vec{\Psi}_{\mathrm{net}}} 
\newcommand{\ball}{\vec{b}} 
\newcommand{\landEvent}{\mathcal{L}} 
\newcommand{\net}{\mathcal{N}} 
\newcommand{\hit}{\mathcal{H}} 

\newcommand{\dynamics}{\vec{f}}
\newcommand{\dynamicsNominal}{\dynamics_{\mathrm{nom}}}

\newcommand{\ValueFunction}{V}
\newcommand{\fullCost}{\mathcal{J}}

\newcommand{\discretization}{\delta} 

\newcommand{\alg}{\emph{bayesILC}}

\newcommand{\sysIdRegressor}{\vec{Y}} 

\newcommand{\fbMat}{\vec{K}} 
\newcommand{\ricMat}{\vec{P}} 
\newcommand{\matOne}{\vec{\Phi}} 
\newcommand{\matTwo}{\vec{\Psi}}
\newcommand{\matThree}{\vec{M}}
\newcommand{\ffVec}{\vec{\ell}} 
\newcommand{\ricVecOne}{\vec{b}} 
\newcommand{\ricVecTwo}{\vec{\nu}} 
\newcommand{\ricScalar}{c}

\newcommand{\param}{\vec{\theta}} 
\newcommand{\dataMat}{\vec{X}} 
\newcommand{\obs}{\vec{y}} 
\newcommand{\mean}{\vec{\mu}} 
\newcommand{\var}{\vec{\Sigma}} 
\newcommand{\noise}{\sigma^{2}} 
\newcommand{\forget}{\lambda} 

\begin{document}
\title{Optimizing the Execution of Dynamic Robot Movements with Learning Control} %

\author{Okan Ko\c c$^{1}$, Guilherme Maeda$^{2}$, Jan Peters$^{1,3}$
\\
{\tt\small \{okan.koc, jan.peters\}@tuebingen.mpg.de, g.maeda@atr.jp}%
\thanks{$^{1}$Max Planck Institute for Intelligent Systems,
        Spemannstr. 38, 72076 Tuebingen, Germany}
\thanks{$^{2}$ATR, Department of Brain Robot Interface,
	Computational Neuroscience Laboratories,
	2-2-2 Hikaridai, Seika-cho, Soraku-gun,
	Kyoto 619-0288, Japan}
\thanks{$^{3}$Technische Universitaet Darmstadt, FG Intelligente Autonome Systeme
        Hochschulstr. 10, 64289 Darmstadt, Germany}
}

\maketitle

\begin{abstract}
  
  High-speed robotics typically involves fast dynamic trajectories
  with large accelerations. Kinematic optimization using compact
  representations can lead to an efficient online computation of these
  dynamic movements, however successful execution requires accurate
  models or aggressive tracking with high-gain feedback. Learning to
  track such references in a safe and reliable way, whenever accurate
  models are not available, is an open problem. Stability issues
  surrounding the learning performance, in the iteration domain, can
  prevent the successful implementation of model-based learning
  approaches. To this end, we propose a new adaptive and
  \emph{cautious} Iterative Learning Control (ILC) algorithm where the
  stability of the control updates is analyzed probabilistically: the
  covariance estimates of the adapted local linear models are used to
  increase the probability of update monotonicity, exercising caution
  during learning.  The resulting learning controller can be
  implemented efficiently using a recursive approach. We evaluate it
  extensively in simulations as well as in our robot table tennis
  setup for tracking dynamic hitting movements. Testing with two seven
  degree of freedom anthropomorphic robot arms, we show improved and
  more stable tracking performance over high-gain PD-control,
  model-free ILC (simple PD feedback type) and model-based ILC without
  cautious adaptation. 
  

  
\end{abstract}

\IEEEpeerreviewmaketitle

\section{Introduction}\label{intro}

\begin{figure}[b!]
\center
\includegraphics[scale=0.18]{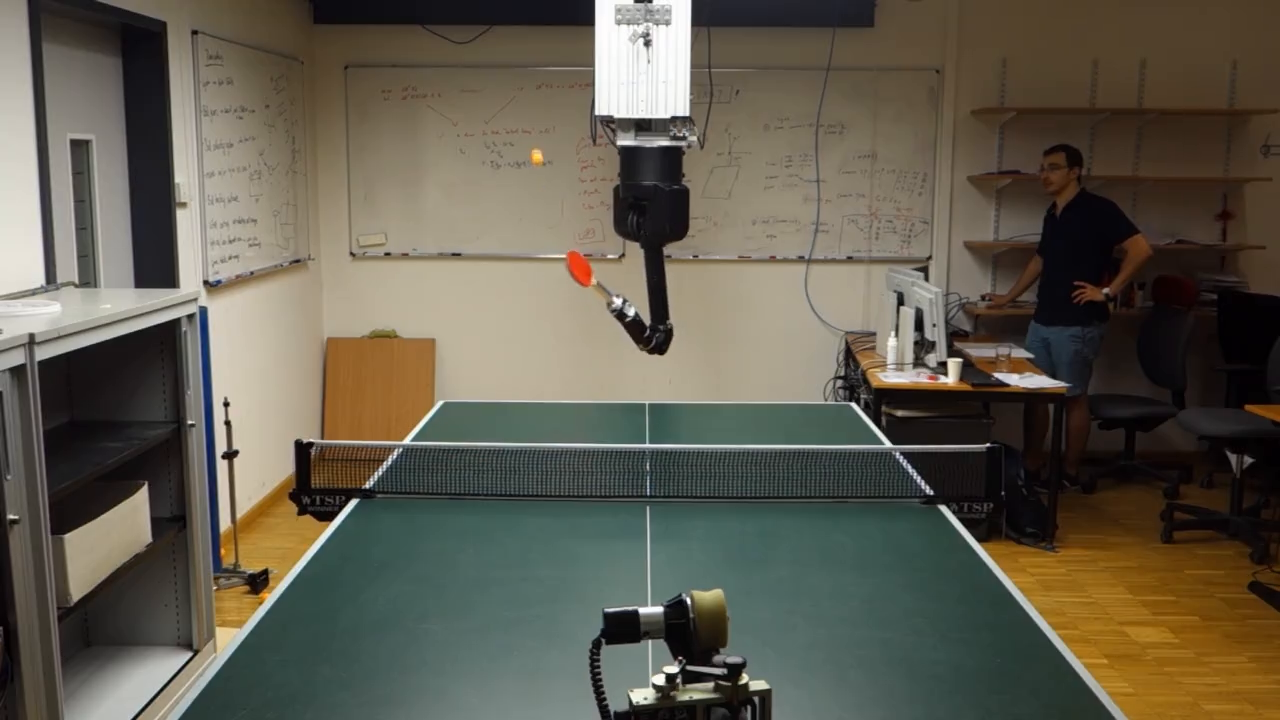}			
\caption{Our robot table tennis platform where a seven degree of freedom Barrett WAM arm is shown facing a ball-launcher. The ball is tracked using four cameras on the ceiling. Whenever a ball is approaching the robot, reference trajectories are computed online in order to return the ball to a desired location on the opponent's court. Such trajectories can be optimized on the kinematics level~\cite{Koc18}, however it is hard to execute them accurately without having access to accurate dynamics models. Iterative Learning Control, using inaccurate models, can still lead to an efficient approach for learning to track these trajectories.} 
\label{robot}
\end{figure}

Most dynamic tasks in robotics include a \emph{tracking} component,
where the system is controlled to follow a desired reference
trajectory. Robot table tennis~\cite{Muelling13},\cite{Koc18}, in
particular involves the generation of fast dynamic trajectories with
high accelerations. These trajectories can be optimized well on the
kinematics level, but reaching the target state and returning the ball
requires accurate tracking of these hitting movements. Computing the
appropriate control inputs for tracking can be a challenging task,
especially when using cable-driven arms, such as the Barrett WAM shown
in Figure~\ref{robot}, due to mechanical compliance and low bandwidth.


Iterative Learning Control (ILC) is a control theoretic learning framework restricted to tracking (time-varying) reference trajectories\cite{Bristow06}. In ILC, the goal is to improve the tracking performance, reducing the future deviations along the fixed trajectory, and ultimately driving them to the minimum possible. After observing the deviations from the reference trajectory at each iteration, the errors are fed back to the (feedforward) control inputs for the next iteration.
Any available dynamics models can be incorporated easily during these updates, see e.g., \cite{Moore07}, \cite{Amann95}.
ILC has been used successfully in several robotics tasks to improve trajectory tracking performance under unknown repeating disturbances and model mismatch \cite{Bristow06}. 

\begin{figure*}[t!]
	\centering
	\begin{minipage}{.6\textwidth}
		\setlength{\figureheight}{6cm}
		\setlength{\figurewidth}{10cm}
		\input{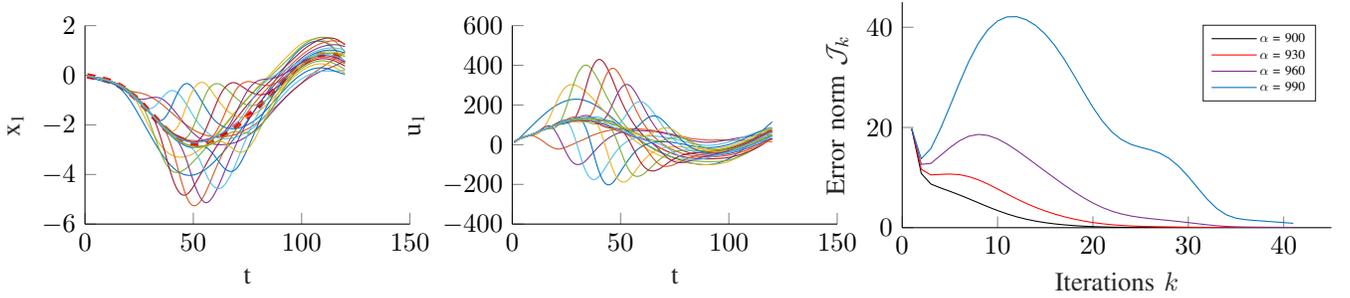}
	\end{minipage}%
	\begin{minipage}{.4\textwidth}
		\setlength{\figureheight}{3cm}
		\setlength{\figurewidth}{6cm}
%
%
\definecolor{mycolor1}{rgb}{0.49412,0.18431,0.55686}%
\definecolor{mycolor2}{rgb}{0.00000,0.44706,0.74118}%
\begin{tikzpicture}

\begin{axis}[%
width=0.951\figurewidth,
height=\figureheight,
at={(0\figurewidth,0\figureheight)},
scale only axis,
xmin=0,
xmax=45,
xlabel style={font=\color{white!15!black}},
xlabel={Iterations $k$},
ymin=0,
ymax=45,
ylabel style={font=\color{white!15!black}},
ylabel={Error norm $\fullCost_k$},
axis background/.style={fill=white},
axis x line*=bottom,
axis y line*=left,
legend style={legend cell align=left, align=left, draw=white!15!black}
]
\addplot [color=black]
  table[row sep=crcr]{%
1	19.7206651861877\\
2	10.8323937663913\\
3	8.71956370373173\\
4	7.94863117893861\\
5	7.26235205236965\\
6	6.5343539369398\\
7	5.75931877136914\\
8	4.95593376913391\\
9	4.16147829378084\\
10	3.41492791732798\\
11	2.74651280894399\\
12	2.17216920070783\\
13	1.69408824868103\\
14	1.30601506899087\\
15	0.997225742466425\\
16	0.754296316419005\\
17	0.564086173207522\\
18	0.416101478229956\\
19	0.302334305040611\\
20	0.216356354555153\\
21	0.152769142589944\\
22	0.106944181624034\\
23	0.0748381925128351\\
24	0.0528944636726581\\
25	0.0381071833565878\\
26	0.0281132579315399\\
27	0.0212007951422667\\
28	0.0162273456101082\\
29	0.0124792957115029\\
30	0.00952853591935\\
31	0.00713121282070961\\
32	0.00516776041122748\\
33	0.00359814810794753\\
34	0.00241673571882346\\
35	0.00160863846864588\\
36	0.00111487126469373\\
37	0.00082922201908001\\
38	0.00064846357378081\\
39	0.000517376454469677\\
40	0.000416961726394727\\
41	0.000339483404406429\\
};
\addlegendentry{$\alpha\text{ = 900}$}

\addplot [color=red]
  table[row sep=crcr]{%
1	19.7206651861877\\
2	11.6734400735538\\
3	10.526231411118\\
4	10.6333698852334\\
5	10.7029173560009\\
6	10.5720217511526\\
7	10.1856130035177\\
8	9.54219775432114\\
9	8.69646128932683\\
10	7.73044875313785\\
11	6.72787015237983\\
12	5.75409115974082\\
13	4.85074097043903\\
14	4.04068618037354\\
15	3.33033680984745\\
16	2.713194768012\\
17	2.17975856630342\\
18	1.72360906960848\\
19	1.34062894554566\\
20	1.02695015995875\\
21	0.777831816803614\\
22	0.586958084586921\\
23	0.445929434892979\\
24	0.344677713066196\\
25	0.272812621011432\\
26	0.221035822839258\\
27	0.182028617246969\\
28	0.150625770925276\\
29	0.123506544804134\\
30	0.0988391059167433\\
31	0.076048749011691\\
32	0.0555537145126281\\
33	0.0383203122983274\\
34	0.0252807834639984\\
35	0.0167422239926768\\
36	0.0119405160269191\\
37	0.0093038400075808\\
38	0.00756712690113281\\
39	0.00621606189657645\\
40	0.00509575998812274\\
41	0.00414617176132247\\
};
\addlegendentry{$\alpha\text{ = 930}$}

\addplot [color=mycolor1]
  table[row sep=crcr]{%
1	19.7206651861877\\
2	12.6374929197352\\
3	12.8197094166631\\
4	14.380374975083\\
5	15.9784264020165\\
6	17.3412698521809\\
7	18.2602895284858\\
8	18.6155863462941\\
9	18.4073587259658\\
10	17.7233666755013\\
11	16.692947100438\\
12	15.4417828378023\\
13	14.0737901887455\\
14	12.6658609166862\\
15	11.2582189264592\\
16	9.86491938632844\\
17	8.50186692391547\\
18	7.19838510762125\\
19	5.99177007257132\\
20	4.91954236171873\\
21	4.0122497261293\\
22	3.28522842544257\\
23	2.73264422862282\\
24	2.32871605298984\\
25	2.03527470973485\\
26	1.81126873189345\\
27	1.62011835574109\\
28	1.43358622665081\\
29	1.23407487478511\\
30	1.0167737921297\\
31	0.790269419072488\\
32	0.573506737181376\\
33	0.388977291783778\\
34	0.254151420785374\\
35	0.173180303870395\\
36	0.132419460558705\\
37	0.110774913793673\\
38	0.0955725006628023\\
39	0.0824363041829172\\
40	0.0701655286153308\\
41	0.0589206510757056\\
};
\addlegendentry{$\alpha\text{ = 960}$}

\addplot [color=mycolor2]
  table[row sep=crcr]{%
1	19.7206651861877\\
2	13.7574602499728\\
3	15.7701437559078\\
4	19.6949471919799\\
5	24.2077985854624\\
6	28.8838408826151\\
7	33.2349145108381\\
8	36.8596561837027\\
9	39.5438672766275\\
10	41.2522108042289\\
11	42.0662340484737\\
12	42.1091709126192\\
13	41.507022709466\\
14	40.345150086015\\
15	38.6341451195746\\
16	36.3630274379693\\
17	33.5842050123589\\
18	30.4417946061032\\
19	27.1578316359817\\
20	24.003032145589\\
21	21.245421937007\\
22	19.0789871482995\\
23	17.56503147077\\
24	16.6172236430799\\
25	16.0324573009256\\
26	15.5451664632639\\
27	14.8824106313115\\
28	13.8195233124536\\
29	12.2416589709764\\
30	10.1942041292607\\
31	7.88927539332209\\
32	5.64921056907356\\
33	3.80061229780332\\
34	2.5535390975214\\
35	1.89477822609452\\
36	1.59762980535747\\
37	1.43545077973032\\
38	1.30053973776484\\
39	1.15899797989922\\
40	1.01262915334032\\
41	0.876400592514094\\
};
\addlegendentry{$\alpha\text{ = 990}$}

\end{axis}
\end{tikzpicture}%
	\end{minipage}
	\caption{Learning performance of ILC, using inaccurate models without incorporating a notion of uncertainty, may not be monotonic in practice. One can observe ripples that move through the trajectory which can cause instability or damage the robot. In simulations we can create this effect easily by increasing the spectral norm of the difference between the nominal and the actual (lifted) dynamics matrices. The desired trajectory for the first state $\state_1$ is shown in dashed red on the left-hand side for a two dimensional linear time invariant system. The second plot shows the ILC feedforward commands for this particular trajectory and state. The third plot shows the Frobenius norm of the trajectory deviations, $\fullCost_k$, plotted over the iterations $k$. The nonmonotonicity of the learning performance is aggravated, as the mismatch scale $\alpha$ controlling the spectral norm of the difference is increased. Increasing $\alpha$ further can prevent even asymptotic stability. The curves were generated by direct inversion of the (lifted) model. Our proposed Bayesian approach, on the other hand, minimizing the expected cost throughout the iterations, uses the posterior over the dynamics model parameters to make more cautious decisions.}
	\label{lin_mc_sim_example}
\end{figure*}

While there have been many impressive applications of reinforcement learning (RL)~\cite{Sutton98} to learn robotic tasks~\cite{Kober08}, RL remains to be computationally and information-theoretically hard in general. Much of control, on the other hand, can be reduced to supervised learning, with the appropriate reference trajectories.
By making good use of existing, albeit imperfect, models and smooth reference trajectories with ILC, learning efficiency in robotics tasks can be improved significantly. 
However, it is rather difficult to ensure a stable learning performance in practice, see Figure~\ref{lin_mc_sim_example} for an illustration.

In this paper, we introduce a new model-based learning approach for tracking a variety of fast, dynamic movements stably, while maintaining learning and computational efficiency.  Stability of the updates, or the probability of update monotonicity, is increased by making use of dynamics model covariance estimates. We refer to this as \emph{caution} throughout the text, and the resulting algorithm is cautious precisely in this sense. A cautious learning control algorithm can hence be defined as one that incorporates a probabilistic notion of stability (in the iteration domain) during decision making, for the control input updates. This property proves to be critical, as we show the learning performance for fast robot table tennis striking movements. The proposed Bayesian approach, using the posterior over the dynamics model parameters, maintains both adaptation and caution in model-based ILC, while being efficient in terms of learning performance and computational complexity.

Our contributions can be stated succinctly as follows: we propose a new adaptive and cautious model-based ILC algorithm, that is implemented efficiently using a recursive formulation. More specifically, the existing model-based recursive ILC approach of Amann et al.~\cite{Amann95}, introduced briefly in Section~\ref{problem_statement}, is extended to include adaptation (by using Linear Bayes Regression on the errors) and caution (or in other terms, robustness to modelling errors, which shows itself as learning stability in the iteration domain). The proposed approach minimizes an expected quadratic cost term over the trajectory deviations, which still yields a closed-form solution, resulting in a cautious yet efficient learning performance. In the closed-form solution, the covariances of the learned local linear models are employed as adaptive regularizers.

The expected cost minimization distinguishes the framework from more conservative min-max approaches, such as the robustly convergent ILC proposed in the literature (using $H_\infty$ and $\mu$-synthesis techniques~\cite{Son16}). Related work in the theory and practice of ILC, as well as some more general applications of learning in robotics tasks, are briefly mentioned in the next subsection. 
Before introducing the expected cost minimization framework in Section~\ref{caution}, we discuss model adaptation in Section~\ref{adaptation} with linear time-varying models and show that Broyden's method~\cite{Nocedal99} can be derived from Linear Bayesian Regression (LBR) as the forgetting factor goes to zero. Thus, the proposed approach belongs to the family of Quasi-Newton ILC methods~\cite{Avrachenkov98}.

The resulting adaptive and cautious ILC algorithm, called $\alg$ is described in Section~\ref{implementation}, and extensions are discussed for additional robustness to nonrepetitive disturbances. Derivations for the recursive and cautious learning control update are left to the Appendix~\ref{derivations}. We evaluate $\alg$ first in extensive simulations in Section~\ref{experiments}, showing that the proposed method is stable, efficient and can outperform other state-of-the-art learning approaches. We then present online learning results on our robot table tennis platform for tracking dynamic hitting movements. Appendix~\ref{hitting} briefly introduces the parameterization of these hitting movements. We discuss the real robot learning results in Section~\ref{conclusion} and conclude with brief mentions of promising future research directions. 





%

\subsection{Related Work}\label{relatedWork}

Since the eighties, there have been many different Iterative Learning Control update laws proposed, with the D-type update law of Arimoto et al.~\cite{Arimoto84} being one of the first. See~\cite{Bristow06} and \cite{Moore07} for reviews and categorization of the different update laws. Theoretically, most ILC updates are \emph{linear repetitive processes} that can be analyzed using 2D-systems analysis~\cite{Rogers07}, i.e., assuming the desired trajectory is fixed and the initial conditions can be reset perfectly, the error over the iterations has a (discrete) dynamics of its own.  Stability of the ILC updates and monotonic convergence in particular can then be studied using dynamical systems theory. These notions also play an important role in the design of practical ILC algorithms. See~\cite{Bristow06}, \cite{Norrloef02} for a discussion and \cite{Longman2000} for insight into convergence and stability issues appearing in an implementation.

Stability issues and the induced oscillations (see Figure~\ref{lin_mc_sim_example} for a simple simulation example) can easily damage the system to be controlled. For instance, joint limits can be exceeded in a robotics application or other task-imposed state constraints can be violated. Such issues complicate the application of ILC in high dimensional robotics problems. In practice, additional complications can occur, such as varying initial conditions, violating the assumptions made in most of the ILC literature. Robustness to varying initial conditions were considered e.g., in \cite{Hillenbrand00}, \cite{Park00}, \cite{Fang03}. For additional robustness to nonrepeating disturbances or noise, a robust feedback controller should be used alongside ILC, see e.g., \cite{Chin04}, \cite{Longman2000}. 



Methods that learn to track (periodic or episodic) trajectories need to compensate for modeling uncertainties and other repetitive disturbances acting on the system to be controlled. However, methods that can efficiently learn the dynamics are model-based (e.g. most of optimization-based ILC \cite{Amann95},\cite{Bristow06}) and at least require knowing the correct signs for the linearized dynamics of the system \cite{Kolter09}, \cite{NguyenTuong11}.

When executing model-based learning algorithms on dynamical systems,
it is essential for stability and safety to incorporate a notion of
model uncertainty. Otherwise the learning algorithms can be
overconfident and quickly go unstable~\cite{Longman2000}. One way to
achieve a more stable performance in ILC is to filter the
high-frequency updates. These robust methods are mostly known as
Q-filtering~\cite{Bristow06} and typically incur a trade-off between
stability and performance: the system will often fail to converge to
the minimal steady-state error. In this paper, we use a different
(probabilistic) approach to increase the stability margins of
model-based ILC that does not incur such a trade-off. To that end, we
expand on the previous work of Amann et al.~\cite{Amann95}, one of the
first model-based ILC approaches introducing an optimal-control based
ILC design. The recursive implementation first introduced in this
paper closely relates to numerically-stable plant-inversion
approaches~\cite{Ghosh99}. We extend the recursive formulation to
include adaptation and caution: adaptation of the model parameter
means and variances are performed at each iteration using Linear Bayes
Regression. The resulting Bayesian approach, minimizing the expected
cost throughout the iterations, uses the posterior over the dynamics
model parameters to make more cautious decisions.




Model adaptation in ILC can be studied in the context of solving nonlinear equations. Tracking a fixed reference perfectly corresponds to solving for the control inputs that drive the deviations to zero. Hence, Quasi-Newton methods such as the Broyden's method~\cite{Nocedal99} and generalized secant method~\cite{Barnes65} were proposed as adaptation methods in the ILC literature to update the plant dynamics. Broyden's method, without having access to the gradients of a black-box function $\vec{f}(\vec{x}) \!=\! \vec{0}$, maintains a Jacobian matrix approximation $\vec{F}$. The matrix $\vec{F}$ is updated at each iteration $k$ in order to satisfy the \emph{secant equation} 
\begin{align}
\vec{f}_{k} - \vec{f}_{k-1} = \vec{F}_k(\vec{x}_{k} - \vec{x}_{k-1}),
\end{align}
\noindent which can then be inverted to yield\footnote{Broyden's method can also directly update the inverse.}
\begin{align}
\vec{x}_{k+1} = \vec{x}_{k} - \vec{F}_k^{\dagger}\vec{f}_{k}.
\end{align}
%
\noindent Convergence under restrictive assumptions have been shown for Broyden's method. For solving systems of nonlinear equations, arguably efficiency rather than stability or monotonic convergence is of importance, and a simple trust-region approach (based on a merit function) suffices to improve stability. We will show how Broyden's method can be seen as a limiting case of Linear Bayesian Regression in Section~\ref{adaptation}. The proposed method thus belongs to the family of Quasi-Newton optimization methods~\cite{Nocedal99}, where the black-box nature of the Quasi-Newton approaches is augmented to include caution during the ILC updates: monotonic convergence, or update stability in the iteration domain, is of paramount importance in robotics tasks.


An application of model-based ILC to reject repeating disturbances was shown in quadrocopter flight~\cite{Schoellig12}, where a constrained convex optimization with imposed control input limits was solved, rather than a direct inversion of the nominal model dynamics. An impressive application of ILC to a robotic surgical task was presented in~\cite{Berg10} utilizing an EM-based ILC update law. ILC was also combined with robust observers to control a heavy-duty hydraulic arm in an excavation task~\cite{Maeda2015Combined}.

Besides ILC, another learning framework that learns inaccurate models for control is model-based Reinforcement Learning. Including variance fully in the decision-making process can result in efficient and stable learning~\cite{Deisenroth11}. However such involved procedures exhibit computational runtime difficulties and have not been implemented in high-dimensional real-time robotics tasks.

\section{Problem Statement and Background}\label{problem_statement}
Most tasks in robotics can be learned more efficiently whenever feasible trajectories are available. Learning-based control approaches can then focus on tracking these trajectories without relying on accurate models. The goal in trajectory tracking is to track a given reference $\traj(t), \, 0 \leq t \leq T \,$, by applying the control inputs $\sysInput(t)$. In dynamic robotic tasks, the references are often in the combined state space of joint positions and velocities $(\vec{q}^{\mathrm{T}}, \dot{\vec{q}}^{\mathrm{T}})^{\mathrm{T}} \in \mathcal{Q} \subset \mathbb{R}^{2n}$, and the control inputs $\sysInput \in \mathcal{U} \subset \mathbb{R}^{m}$ are applied for each joint of the robot, i.e., $m = n$. The reference trajectories in table tennis, for instance, enable the execution of hitting and striking motions, e.g., forehand and backhand strikes. Such trajectories can be generated online with nonlinear constrained optimization~\cite{Koc18}. Finding the right control inputs to track them accurately is the focus of Iterative Learning Control (ILC). 
\vspace{2mm}
\subsubsection{Linearizing an Inaccurate Model} Consider a nonlinear robot dynamics model

\begin{equation}
\begin{aligned}
\ddot{\joint} &= \dynamics(\joint,\dot{\joint},\sysInput),
\end{aligned}
\label{dynamics}
\end{equation}
\noindent e.g., for rigid body dynamics of the form
\begin{equation}
  \begin{aligned}
    \ddot{\joint} &= \vec{M}^{-1}(\joint)\{ \sysInput - \vec{C}(\joint,\dot{\joint})\dot{\joint} - \vec{G}(\joint)\},
  \end{aligned}
  \label{rigidBody}
\end{equation}
%
\noindent where on the right-hand side are the inverse of the inertia matrix $\vec{M}(\joint)$, the Coriolis and centrifugal forces $\vec{C}(\joint,\dot{\joint})\dot{\joint}$, and the vector of gravitational forces $\vec{G}(\joint)$. This nonlinear dynamics model can be linearized around a given joint space trajectory $\traj(t), \ 0 \leq t \leq T$ with nominal inputs $\trjInput(t)$ calculated using the inverse dynamics model~\cite{Spong06}. We then obtain the following linear time-varying (LTV) representation
\begin{equation}
\begin{aligned}
\dot{\error}(t) = \vec{A}(t)\error(t) + \vec{B}(t)\linInput(t) + \linDist(t,\sysInput),
\end{aligned}
\label{LTV}
\end{equation}
\noindent where the state vector is the joint angles and velocities $\state = [\joint^{\intercal},\dot{\joint}^{\intercal}]^{\mathrm{T}}$, the state error is denoted as $\error(t) = \state(t) - \traj(t)$, the deviations from the nominal inputs are $\linInput(t) = \sysInput(t) - \trjInput(t)$ and the continuous time-varying matrices are
\begin{equation}
\begin{aligned}
\vec{A}(t) = \at{\frac{\partial{\dynamics}}{\partial{\state}}}{(\traj(t),\trjInput(t))}, 
\vec{B}(t) = \at{\frac{\partial{\dynamics}}{\partial{\sysInput}}}{(\traj(t),\trjInput(t))}.
\end{aligned}
\label{LTVmatrices}
\end{equation}
%
\noindent In the error dynamics \eqref{LTV}, the additional (unknown) term $\linDist(t,\sysInput)$ accounts for the disturbances and the effects of the linearization (i.e., higher order terms). We can discretize (\ref{LTV}-\ref{LTVmatrices}) with step size $\discretization$, $N = T/\discretization$ and step index $j = 1, \ldots, N$ to get the following discrete-time linear system
\begin{equation}
\begin{aligned}
\error_{j+1} = \vec{A}_j\error_j + \vec{B}_j\linInput_j + \linDist_{j+1},
\end{aligned}
\label{discreteLTV}
\end{equation}
\noindent where the matrices $\vec{A}_j, \vec{B}_j$ are the discretizations of \eqref{LTVmatrices}. Conventional (discrete) ILC algorithms learn to compensate for the errors incurred along the trajectory by updating the control inputs $\linInput_j$ iteratively.

Whenever we refer to the outcome of a particular iteration $k$, we will use the first subindex for iterations and the second subindex will be used to denote the (discrete) time step, i.e., the vectors $\error_{k,j} \in \mathbb{R}^{2n}, \ \linInput_{k,j} \in \mathbb{R}^{m}$ denote the deviations and control input compensations at the time step $j$ during iteration $k$, respectively. The control commands applied at iteration $k+1$ as
\begin{equation}
\begin{aligned}
\sysInput_{k+1,j} = \sysInput_{k,j} + \linInput_{k,j},
\end{aligned}
\end{equation}
\noindent are computed using the deviations $\error_{k,j}$ at iteration $k$.  

\vspace{2mm}
\subsubsection{Recursive Norm-Optimal ILC} 

Norm-optimal ILC uses the discrete LTV model in \eqref{discreteLTV} to minimize the next iteration errors, where the computed control inputs are optimal with respect to some vector norm. These approaches based on optimality criteria can learn efficiently by taking advantage of the inaccurate models. Batch methods that compute the next iteration compensations stack the model matrices together to compute (a possibly weighted and dampened) pseudoinverse of this block lower-diagonal matrix. As an alternative, some methods use convex programming to compute these optimal compensations under additional constraints. 

The condition of this \emph{lifted} model matrix typically grows exponentially with the horizon size $N$ and computing the pseudoinverse stably becomes very difficult. Downsampling trajectories restores the condition number and a stable inversion becomes much more manageable, at the cost of reduced tracking performance. As a better alternative, optimization-based approaches, depending on the particular optimizer, may avoid computing the pseudoinverse. However such approaches can still be computationally intensive, and may not be suitable for online learning. 

As an alternative, the authors in~\cite{Amann95} have shown that the direct batch inversion of the lifted model matrix can be avoided by recursively computing the ILC compensations (in one pass) using the Linear Quadratic Regulator (LQR) for disturbance rejection~\cite{AndersonMoore}. After estimating the disturbances $\linDist_{j+1}$ at the $k'$th trial, the optimal control problem for tracking a desired trajectory can be written as
\begin{equation}
\begin{aligned}
\min_{\linInput} & \, \sum_{j = 1}^{N} \, \error^{\mathrm{T}}_{k+1,j}\vec{Q}_j\error_{k+1,j} + \linInput^{\mathrm{T}}_{k,j}\vec{R}_j\linInput_{k,j}, \\
\textrm{s.t. \ } & \error_{k+1,j+1} = \vec{A}_j\error_{k+1,j} + \vec{B}_j\sysInput_{k+1,j} + \linDist_{j+1}.
\end{aligned}
\label{lqr-deterministic-cost}
\end{equation}
Reduction of the ILC problem to the known LQR solution has not attracted much attention however from the control and learning communities, since it was not clear how to study stability and convergence in this formulation.
%

\section{Model Adaptation}\label{adaptation}
Whenever there is model-mismatch, the (linearized) models cannot be assumed to hold accurately around the reference trajectory. There is hence a risk that the learning process described in the previous subsections will not be stable. As a remedy, in this section we propose a natural Bayesian adaptation of model matrices with Linear Bayesian Regression (LBR) and discuss different alternatives in the context of robotics. 
\vspace{2mm}
\subsection{Recursive Estimation of Model Matrices}\label{adaptiveILC}
The observed deviations from the trajectory at iteration $k$, $\error_{k,j}$, can be used to update the discrete-time LTV model matrices $\vec{A}_{k,j}, \vec{B}_{k,j}$ that describe the nonlinear dynamics around the trajectory, to first order. Instead of estimating all the parameters together in a costly estimation procedure, the model matrices $\vec{A}_{k,j}, \vec{B}_{k,j}$ can rather be updated separately for each $j = 1, \ldots, N$, given the smoothened errors $\hat{\error}_{k,j}$
\begin{equation}
\begin{aligned}
\hat{\error}_{k,j+1} = \vec{A}_{k,j}\hat{\error}_{k,j} + \vec{B}_{k,j}\sysInput_{k,j} + \linDist_{j+1},
\end{aligned}
\label{discr_model}
\end{equation}
\noindent which can be rewritten using the Kronecker product and the vectorization operator as follows
\begin{equation}
\begin{aligned}
&\hat{\error}_{k,j+1} - \hat{\error}_{k-1,j+1} \approx \dataMat_{k,j}\vect{[\vec{A}_{k,j}, \vec{B}_{k,j}]}, \\
&\dataMat_{k,j} = \vect{[\hat{\error}_{k,j} - \hat{\error}_{k-1,j}, \linInput_{k,j}]}^{\mathrm{T}} \otimes \vec{I}. \\
\end{aligned}
\end{equation}
If we incorporate the belief (including the uncertainty) about the linear dynamics models as Gaussian priors in LBR
\begin{equation}
\begin{aligned}
&\param_{k,j} = \vect{[\vec{A}_{k,j}, \vec{B}_{k,j}]}, \\
&\obs_{k,j} = \hat{\error}_{k,j+1} - \hat{\error}_{k-1,j+1}, \\
&\rho(\param_{k,j}|\vec{y}_{k,j}) \propto \rho(\obs_{k,j}|\param_{k,j})\rho(\param_{k,j}), \\
&\rho(\param_{k,j}) = \mathcal{N}(\param_{k,j}|\mean_{k,j},\var_{k,j}),
\end{aligned}
\label{model-params}
\end{equation}
\noindent with a Gaussian likelihood function 
\begin{equation}
\begin{aligned}
\rho(\obs_{k,j}|\param_{k,j}) = \mathcal{N}(\obs_{k,j}|\dataMat_{k,j}\param_{k,j},\noise\vec{I}),
\end{aligned}
\end{equation}
\noindent the models parameter means $\mean_{k,j}$ and variances $\var_{k,j}$ can be updated as
%
%
\begin{equation}
\begin{aligned}
\var_{k,j} &= (\tfrac{1}{\noise}\dataMat^{\mathrm{T}}_{k,j}\dataMat_{k,j} + \var_{k-1,j}^{-1})^{-1}, \\
\mean_{k,j} &= \var_{k,j}\left(\var_{k-1,j}^{-1}\mean_{k-1,j} + \tfrac{1}{\noise}\dataMat^{\mathrm{T}}_{k,j}\obs_{k,j}\right).
\end{aligned}
\label{update_model_matrices}
\end{equation}
Smoothened position and velocity error estimates can be obtained, for example, using a zero-phase Butterworth filter.
\vspace{2mm}
\subsubsection{Relation to Broyden's method} 
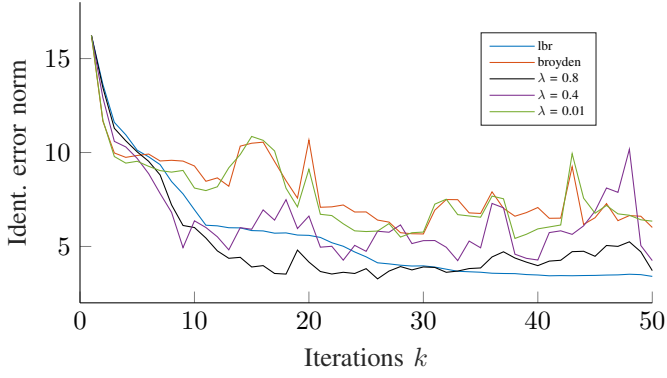
\begin{figure}[t!]
	\setlength{\figureheight}{4cm}
	\setlength{\figurewidth}{8cm}
%
%
\definecolor{mycolor1}{rgb}{0.00000,0.44700,0.74100}%
\definecolor{mycolor2}{rgb}{0.85000,0.32500,0.09800}%
\definecolor{mycolor3}{rgb}{0.49400,0.18400,0.55600}%
\definecolor{mycolor4}{rgb}{0.46600,0.67400,0.18800}%
\begin{tikzpicture}

\begin{axis}[%
width=0.951\figurewidth,
height=\figureheight,
at={(0\figurewidth,0\figureheight)},
scale only axis,
xmin=0,
xmax=50,
xlabel style={font=\color{white!15!black}},
xlabel={Iterations $k$},
ymin=2,
ymax=18,
ylabel style={font=\color{white!15!black}},
ylabel={Ident. error norm},
axis background/.style={fill=white},
axis x line*=bottom,
axis y line*=left,
legend style={legend cell align=left, align=left, draw=white!15!black}
]
\addplot [color=mycolor1]
  table[row sep=crcr]{%
1	16.2314312294422\\
2	13.6242587673348\\
3	11.602882125872\\
4	10.9433270246539\\
5	10.1204054095658\\
6	9.7710828610034\\
7	9.35508371488628\\
8	8.45467152893961\\
9	7.79959966926091\\
10	6.95308338021904\\
11	6.13068552400765\\
12	6.09768660682918\\
13	5.99725995528006\\
14	5.98904904507026\\
15	5.84609425069673\\
16	5.82757828724041\\
17	5.70704721413755\\
18	5.72153833641295\\
19	5.60797055573769\\
20	5.59123268511624\\
21	5.48103295797896\\
22	5.20033471067856\\
23	5.01216844606361\\
24	4.70736074267747\\
25	4.44810150375377\\
26	4.13056503865338\\
27	4.07315639670936\\
28	4.00291317034963\\
29	3.95926225427569\\
30	3.96798983441193\\
31	3.90280198438205\\
32	3.79163628663914\\
33	3.68052253886683\\
34	3.6541716117221\\
35	3.62886303455136\\
36	3.57551643148245\\
37	3.56160767706872\\
38	3.55499783644716\\
39	3.50504542182642\\
40	3.48227273198085\\
41	3.4396000121617\\
42	3.44999282308742\\
43	3.44236135298004\\
44	3.45355232685012\\
45	3.46004335050279\\
46	3.47811288030371\\
47	3.48792746124257\\
48	3.52259119836361\\
49	3.50068174902726\\
50	3.40793557573624\\
};
\addlegendentry{lbr}

\addplot [color=mycolor2]
  table[row sep=crcr]{%
1	16.2314312294422\\
2	11.6661889918291\\
3	9.9761774025067\\
4	9.74229251162807\\
5	9.84169648802074\\
6	9.91811261410222\\
7	9.55005854771808\\
8	9.58634120407064\\
9	9.54410988886223\\
10	9.28251470189992\\
11	8.46867380391953\\
12	8.64944084323573\\
13	8.20540472234268\\
14	10.3402533347313\\
15	10.4942707052029\\
16	10.5507016681319\\
17	9.51923627614612\\
18	8.50178989206556\\
19	7.57300745445157\\
20	10.6633720401559\\
21	7.08588640548992\\
22	7.10886833798717\\
23	7.20752013260347\\
24	6.834117019001\\
25	6.83800586734615\\
26	6.40485395252475\\
27	6.29668522906072\\
28	5.73199947073838\\
29	5.6774295597574\\
30	5.66572052249122\\
31	6.93328455175267\\
32	7.49740851934521\\
33	7.49055590880402\\
34	6.7877328324459\\
35	6.75612587286266\\
36	7.90673174080349\\
37	7.02902842459854\\
38	6.61066284123789\\
39	6.80207051413486\\
40	7.07300319544695\\
41	6.49902601010058\\
42	6.50688125798734\\
43	9.24350759583811\\
44	6.17285356800001\\
45	6.52945223508801\\
46	7.28185711600665\\
47	6.37436226025931\\
48	6.64430687242108\\
49	6.60814039361988\\
50	6.01255878670183\\
};
\addlegendentry{broyden}

\addplot [color=black]
  table[row sep=crcr]{%
1	16.2314312294422\\
2	13.4536418569624\\
3	11.2904131318193\\
4	10.6239835443042\\
5	10.0332273649655\\
6	9.53154702968534\\
7	8.80630400937096\\
8	7.21181668466108\\
9	6.11647105526029\\
10	6.00807289410301\\
11	5.44839301885479\\
12	4.76544863239323\\
13	4.3707396693666\\
14	4.42168325073797\\
15	3.90840886598061\\
16	3.98056829291919\\
17	3.56327667644636\\
18	3.52953667018072\\
19	4.80777925664572\\
20	4.17112158781414\\
21	3.66723063955235\\
22	3.53425267705089\\
23	3.62609862668902\\
24	3.56231093625478\\
25	3.81956157298309\\
26	3.27569617802117\\
27	3.69086859811834\\
28	3.93006365426021\\
29	3.75252391483026\\
30	3.90981342909595\\
31	3.88418629970858\\
32	3.62502412669001\\
33	3.69077089162647\\
34	3.82447564238192\\
35	3.85735067975047\\
36	4.44273436338075\\
37	4.716608173534\\
38	4.38584667478806\\
39	4.16726261626887\\
40	3.98037483104256\\
41	4.220891708008\\
42	4.2657853042273\\
43	4.72353488543429\\
44	4.74455832932786\\
45	4.47692532953077\\
46	5.05180247994769\\
47	5.00511014565048\\
48	5.24751499058558\\
49	4.71518899816458\\
50	3.71450689070175\\
};
\addlegendentry{$\lambda\text{ = 0.8}$}

\addplot [color=mycolor3]
  table[row sep=crcr]{%
1	16.2314312294422\\
2	12.9174011558829\\
3	10.593225377199\\
4	10.2976363229247\\
5	9.68452209954468\\
6	8.87356059056989\\
7	7.81864102327773\\
8	6.81431052423203\\
9	4.93970126558427\\
10	6.36545166993931\\
11	6.00951613367456\\
12	5.51628559174739\\
13	4.82128378858702\\
14	6.00072173653612\\
15	5.91586072636465\\
16	6.94689066773809\\
17	6.40180070488921\\
18	7.4901231368361\\
19	5.95406714734905\\
20	6.61730682777858\\
21	4.96155626453834\\
22	5.0102085108317\\
23	4.2708534696473\\
24	5.05888076192871\\
25	4.74077762035237\\
26	5.81029941154989\\
27	5.75730504158266\\
28	6.14711096949429\\
29	5.15464504755378\\
30	5.30985444228497\\
31	5.31616980668431\\
32	4.95754321022332\\
33	4.24643437915949\\
34	5.28716605421096\\
35	4.92878668938928\\
36	7.28951440222746\\
37	7.05593875031258\\
38	4.58895869810855\\
39	4.36838705744967\\
40	4.27860747203193\\
41	5.73821824871299\\
42	5.83335235277186\\
43	5.64438606533854\\
44	6.09675045931978\\
45	6.88458309292151\\
46	8.11156352297673\\
47	7.88016925401761\\
48	10.1725422293273\\
49	5.05136808133326\\
50	4.2492645873112\\
};
\addlegendentry{$\lambda\text{ = 0.4}$}

\addplot [color=mycolor4]
  table[row sep=crcr]{%
1	16.2314312294422\\
2	11.7170698979524\\
3	9.78602623156983\\
4	9.43627426652048\\
5	9.53366869929624\\
6	9.26427574863679\\
7	9.03717281113992\\
8	8.95625518305289\\
9	9.04649259046465\\
10	8.10900258658018\\
11	7.97379048334904\\
12	8.17688407057066\\
13	9.18783134561121\\
14	9.8931255937095\\
15	10.8550418003346\\
16	10.6486618360849\\
17	10.0850062688398\\
18	8.10176119919748\\
19	7.10576228339325\\
20	9.11014316260661\\
21	6.7124038394555\\
22	6.64425365727908\\
23	6.20719713004944\\
24	5.83103162780902\\
25	5.79424850388498\\
26	5.81945410091309\\
27	6.21819069794667\\
28	5.49982371440816\\
29	5.73227930879715\\
30	5.76553024468197\\
31	7.25279284064516\\
32	7.51468199037842\\
33	6.69418747361553\\
34	6.62393239397247\\
35	6.55659923907927\\
36	7.68166637793226\\
37	7.53823423920194\\
38	5.4283553823212\\
39	5.64863856179843\\
40	5.93594689528931\\
41	6.03902947122612\\
42	6.14174616246199\\
43	9.93726619703267\\
44	7.56972591684308\\
45	6.76898123305534\\
46	7.18359575014588\\
47	6.73519686786624\\
48	6.66257680331412\\
49	6.42012425370321\\
50	6.34706315927335\\
};
\addlegendentry{$\lambda\text{ = 0.01}$}

\end{axis}
\end{tikzpicture}%
	\caption{Broyden's method~\cite{Nocedal99}, which can be considered as an adaptation framework within ILC, is a limiting case of Linear Bayesian Regression (LBR). As the forgetting factor $\forget$ of an exponentially weighted LBR model goes to zero, LBR transitions to Broyden's method. Broyden's method is very sensitive to noise and adapts very aggressively. Throughout the paper, we discuss and evaluate several adaptation laws, that are less sensitive to noise but are still flexible. The Figure shows the evolution of the identification error norm for an unknown linear time-varying system. The Frobenius norm of the difference between the adapted model matrices ($\vec{A}_{k,j}$ and $\vec{B}_{k,j}$) and the actual (fixed) matrices (denoted as identification error norm) are plotted for each iteration $k = 1, \ldots, 50$.}
	\label{lbr_to_broyden}
\end{figure}

Broyden's method~\cite{Nocedal99} can be seen as a limiting case of LBR. The mean estimates in \eqref{update_model_matrices} are also the solutions of the following linear ridge regression problem
\begin{equation}
\begin{aligned}
\min_{\param} & \, \tfrac{1}{\noise}\|\obs_{k,j} - \dataMat_{k,j}\param\|_2^2 + (\param - \param_{k,j})\var_{k,j}^{-1}(\param - \param_{k,j}),
\end{aligned}
\end{equation}
and as $\noise \to 0$ we get the (weighted) Broyden's update for one iteration, which, written in vectorized form, is solving independently for every time step
\begin{align}
\min_{\param} & \, (\param - \param_{k,j})\var_{k,j}^{-1}(\param - \param_{k,j}), \\
\textrm{s.t. \ } & \obs_{k,j} = \dataMat_{k,j}\param. \label{secant_rule}
\end{align}
Broyden's method is too sensitive to the sensor noise in robotics tasks as it satisfies the secant rule \eqref{secant_rule} exactly. On the other hand, LBR in \eqref{update_model_matrices} for fixed noise parameter $\noise$, is using \emph{all} of the past iteration data equally. The norm of the covariance decreases monotonically in each update. For unknown dynamic systems that are highly nonlinear but smooth, to prevent premature shrinking of the covariance matrix, a better alternative is to set an \emph{exponential weighting} in the adaptation. For a fixed forgetting factor $\forget \in [0,1]$, the update in \eqref{update_model_matrices} becomes
\begin{equation}
\begin{aligned}
\var_{k,j} &= (\tfrac{1}{\noise}\dataMat^{\mathrm{T}}_{k,j}\dataMat_{k,j} + \forget\var_{k-1,j}^{-1})^{-1}, \\
\mean_{k,j} &= \forget\var_{k,j}\var_{k-1,j}^{-1}\mean_{k-1,j} + \tfrac{1}{\noise}\var_{k,j}\dataMat^{\mathrm{T}}_{k,j}\obs_{k,j}.
\end{aligned}
\label{lbr_with_forgetting}
\end{equation}
\noindent The forgetting factor $\forget$ is used to perform exponential weighting of the previous iteration data. As $\forget \to 0$, we get the (unweighted) Broyden's method\footnote{Unlike the case where $\noise \to 0$, this equivalence is valid for all the subsequent iterations as well. It can be seen more easily from the filter form of \eqref{lbr_with_forgetting}.}, and as $\forget \to 1$, \eqref{lbr_with_forgetting} reduces to \eqref{update_model_matrices}. Hence, our proposed adaptation law \eqref{lbr_with_forgetting} can be embedded within a one-parameter family of Quasi-Newton ILC methods, where the forgetting factor parameter trades-off adaptation flexibility and robustness to noise. At the one end of the spectrum, Broyden's method adapts flexibly and aggressively to the latest data at the cost of being very sensitive to noise. This can be alleviated with a judicious choice of the forgetting factor. See Figure~\ref{lbr_to_broyden} for an illustration.
%
\subsection{Imposing structure} The structure in the forward dynamics model \eqref{rigidBody} is not considered in the update rule \eqref{lbr_with_forgetting}: any change in the control inputs in this model directly affects the instantaneous joint accelerations, and only indirectly the joint velocities in the future time steps. By differentiating the smoothened joint velocities, one can instead impose the following regression model 
\begin{equation}
\begin{aligned}
\ddot{\vec{q}}_{k,j} - \ddot{\vec{q}}_{k-1,j} \approx \vec{A}_{k}(\discretization j)\error_{k,j} + \vec{B}_{k}(\discretization j)\linInput_{k,j},
\end{aligned}
\label{cts_model}
\end{equation}
\noindent where we dropped the hat for notational simplicity. The \emph{continuous} model matrices $\vec{A}_{k}(\discretization j), \vec{B}_{k}(\discretization j)$ are members of a reduced parameter space, i.e., $\vec{A}_{k}(\discretization j) \in \mathbb{R}^{n \times 2n}, \vec{B}_{k}(\discretization j) \in \mathbb{R}^{n \times m}$, $j = 1, \ldots, N$. After regressing on the continuous model matrices
as in \eqref{update_model_matrices}, they can be discretized (as discussed before) to form the discrete-time model parameter means $\vec{A}_{k,j} \in \mathbb{R}^{2n \times 2n}, \vec{B}_{k,j} \in \mathbb{R}^{2n \times m}$ and covariances $\var_{k,j}$.

As an alternative, note that the rigid body dynamics \eqref{dynamics} is parameterized by the link masses, three link center of mass values and six inertia parameters. A total of ten parameters are used for each link to fully parameterize the inverse dynamics model
\begin{equation}
\begin{aligned}
\sysInput &= \vec{M}(\joint;\param)\ddot{\vec{q}} + \vec{C}(\joint,\dot{\joint};\param)\dot{\joint} + \vec{G}(\joint;\param),
\end{aligned}
\end{equation}
\noindent which can be stacked for each $j = 1, \ldots, N$ to form the regression model
\begin{equation}
\begin{aligned}
\vec{U}_k &\approx \sysIdRegressor(\vec{Q}_k^{(0)},\vec{Q}_k^{(1)},\vec{Q}_k^{(2)})\param_k, \\
\vec{U}_k &= \left(\sysInput_{k,1}^{\mathrm{T}}, \sysInput_{k,2}^{\mathrm{T}}, \ldots, \sysInput_{k,N}^{\mathrm{T}}\right)^{\mathrm{T}}, \\
\vec{Q}_k^{(l)} &= \left(\joint_{k,1}^{(l)\mathrm{T}}, \joint_{k,2}^{(l)\mathrm{T}}, \ldots, \joint_{k,N}^{(l)\mathrm{T}}\right)^{\mathrm{T}}, \, l = 0, 1, 2,
\end{aligned}
\label{sys_id}
\end{equation}
where $\param _k\in \mathbb{R}^{10n}$ appears \emph{linearly}. The index $l$ denotes the degree of the derivatives of the smoothened joint angles, i.e., $l = 0, 1, 2$ are used to denote the joint position, velocity and acceleration estimates in \eqref{sys_id}, respectively. Based on these joint estimates, only the link parameters are updated with LBR as in \eqref{update_model_matrices}. The forward dynamics model\footnote{The forward dynamics model \eqref{dynamics}, unlike the inverse dynamics \eqref{sys_id}, depends \emph{nonlinearly} on the link parameters.} \eqref{dynamics} can then be used to sample the means and variances of the continuous LTV matrices, e.g., using Monte Carlo sampling. Discretization as discussed above converts the continuous-time model parameter means and variances into their discrete-time form. An advantage of this approach is to \emph{compress} learning to a lower dimensional space, reducing the variance of the updates at the cost of an introduced bias. Moreover, since the link parameters are invariant throughout the iterations, such an update avoids the flexible yet independent adaptation of the model matrices for each $j$, and the necessity of introducing a forgetting factor.

\section{Cautious Learning Control}\label{caution}
The posterior model covariances $\var_{k,j}$ can be used to make more \emph{cautious} decisions within a stochastic control framework. The uncertainty of the model parameters can be seen as a \emph{multiplicative} noise model and the ILC optimality criterion $\eqref{lqr-deterministic-cost}$ can be extended to include expectations over them. The multiplicative noise model, unlike the additive noise case, does not lead to \emph{certainty-equivalence}: the covariance estimates are incorporated in the decision rule.  
To see how the expected cost minimization leads to caution, note that
\begin{align}
\mathbb{P}(\error^{\mathrm{T}}_{k+1,j}\vec{Q}_j\error_{k+1,j} \geq \hat{\error}^{\mathrm{T}}_{k,j}\vec{Q}_j\hat{\error}_{k,j}) \leq \frac{\mathbb{E}[\error^{\mathrm{T}}_{k+1,j}\vec{Q}_j\error_{k+1,j}]}{\hat{\error}^{\mathrm{T}}_{k,j}\vec{Q}_j\hat{\error}_{k,j}},
\end{align}
\noindent which follows from Markov's inequality. Minimizing the upper bound forces the probability of nonmonotonicity to be low as well. 

\vspace{2mm}
\subsubsection{Expected Cost Minimization}
For the expected cost case, where the expectation is taken over the \emph{random variables} $\vec{A}_{k,j}$ and $\vec{B}_{k,j}$, for each $j$, the optimality criterion
\begin{equation}
\begin{aligned}
\min_{\linInput} & \, \sum_{j = 1}^{N} \mathbb{E}_{\vec{A}_{k,j}, \vec{B}_{k,j}}[\error^{\mathrm{T}}_{k+1,j}\vec{Q}_j\error_{k+1,j} \!+\! \linInput^{\mathrm{T}}_{k,j}\vec{R}_j\linInput_{k,j}], \\
\textrm{s.t. \ } & \error_{k+1,j+1} = \vec{A}_{k,j}\error_{k+1,j} + \vec{B}_{k,j}\sysInput_{k+1,j} + \linDist_{j+1},
\end{aligned}
\end{equation}
\noindent can be solved recursively using dynamic programming~\cite{Kendrick81}
\begin{equation}
\begin{aligned}
\linInput_{k,j} &= \fbMat_{k,j}\error_{k+1,j} - \matOne_{k,j}^{-1}\ffVec_{k,j}, \\
\fbMat_{k,j} &= -\matOne_{k,j}^{-1}\matTwo_{k,j}, \\
\matOne_{k,j} &= \mathbb{E}_{\vec{B}_{k,j}}[\vec{B}_{k,j}^{\mathrm{T}}\ricMat_{k,j+1}\vec{B}_{k,j}] + \vec{R}_{j}, \\
\matTwo_{k,j} &= \mathbb{E}_{\vec{A}_{k,j}, \vec{B}_{k,j}}[\vec{B}_{k,j}^{\mathrm{T}}\ricMat_{k,j+1}\vec{A}_{k,j}], \\
\ffVec_{k,j} &=  \mathbb{E}_{\vec{B}_{k,j}}[\vec{B}_{k,j}^{\mathrm{T}}\ricMat_{k,j+1}(\vec{B}_{k,j}\sysInput_{k,j} \!+\! \linDist_{j+1}) \!+\! \vec{B}_{k,j}^{\mathrm{T}}\ricVecOne_{k,j+1}], 
\end{aligned}
\label{recursive_ilc_cautious}
\end{equation}
\noindent where $\ricVecOne_{k,j}$ and the Ricatti matrices $\ricMat_{k,j}$ evolve backwards according to
\begin{equation}
\begin{aligned}
\ricMat_{k,j} &= \vec{Q}_j + \matThree_{k,j} - \matTwo_{k,j}^{\mathrm{T}}\matOne_{k,j}^{-1}\matTwo_{k,j}, \\
\matThree_{k,j} &= \mathbb{E}_{\vec{A}_{k,j}}[\vec{A}_{k,j}^{\mathrm{T}}\ricMat_{k,j+1}\vec{A}_{k,j}], \\
\ricVecOne_{k,j} &= \mathbb{E}_{\vec{A}_{k,j}, \vec{B}_{k,j}}[\bar{\vec{A}}_{k,j}^{\mathrm{T}}(\ricVecOne_{k,j+1} \!+\! \ricMat_{k,j+1}(\vec{B}_{k,j}\sysInput_{k,j} \!+\! \linDist_{j+1}))], 
\end{aligned}
\end{equation}
\noindent starting from $\ricMat_{k,N} = \vec{Q}_N$ and $\vec{b}_{k,N} = \vec{0}$. The random closed loop system dynamics is given by the matrices
\begin{equation}
\begin{aligned}
\bar{\vec{A}}_{k,j} &= \vec{A}_{k,j} + \vec{B}_{k,j}\fbMat_{k,j}.
\end{aligned}
\phantom{\hspace{6cm}}
\end{equation}
%

By a direct comparison to the LQR solution to $\eqref{lqr-deterministic-cost}$, it can be seen
that the control input compensations $\linInput_{k,j}$ in $\eqref{recursive_ilc_cautious}$ are computed similarly, with the appropriate expectations added. The ILC update is decomposed into two components: a current-iteration feedback term $\sysInput_{\mathrm{fb}} = \fbMat_{k,j}\error_{k+1,j}$ calculated using the iteration dependent Riccati equations and a feedforward, purely predictive term $\sysInput_{\mathrm{ff}} = - \matOne_{k,j}^{-1}\ffVec_{k,j}$, solved backwards for each $j = 1, \ldots, N$. The feedforward terms are responsible for compensating for the estimated \emph{random} disturbances $\vec{d}_{j}$, calculated using \eqref{discr_model}.  

Cautious update $\eqref{recursive_ilc_cautious}$ can be implemented without explicitly calculating the disturbances. If the disturbances are taken as random variables defined via the filtered errors $\hat{\error}_{k,j}$ of the last iteration
\begin{equation}
\begin{aligned}
\linDist_{j+1} &= \hat{\error}_{k,j+1} - \vec{A}_{k,j}\hat{\error}_{k,j} - \vec{B}_{k,j}\sysInput_{k,j},
\end{aligned}
\end{equation}
\noindent the recursion can be simplified by introducing 
\begin{equation}
\ricVecTwo_{k,j} = \ricVecOne_{k,j} + \ricMat_{k,j}\error_{k,j}.
\end{equation}
The feedforward and feedback compensations $\linInput_{k,j}$ can then directly be computed as
\begin{equation}
\begin{aligned}
\linInput_{k,j} &= \fbMat_{k,j}(\error_{k+1,j} \,\textrm{--}\, \error_{k,j}) \,\textrm{--}\, \matOne_{k,j}\mathbb{E}_{\vec{B}_{k,j}}[\vec{B}_{k,j}^{\mathrm{T}}\ricVecTwo_{k,j+1}], \\
\ricVecTwo_{k,j} &= \mathbb{E}_{\vec{A}_{k,j}, \vec{B}_{k,j}}[\bar{\vec{A}}_{k,j}^{\mathrm{T}}\ricVecTwo_{k,j+1}] + \vec{Q}_j\error_{k,j}.
\end{aligned}
\label{recursive_implementation}
\end{equation}
See Appendix~\ref{derivations} for a detailed derivation. Equation \eqref{recursive_implementation} is easier to implement, since the disturbances do not need to be estimated explicitly. The compensations $\linInput_{k,j}$ are added to the total control inputs applied at iteration $k$. In an adaptive implementation, the feedback components of the update, $\fbMat_{k,j}(\error_{k+1,j} - \error_{k,j})$, does not completely subtract the previous feedback controls $\fbMat_{k-1,j}\error_{k,j}$ from the total control inputs, as the feedback matrices are also adapted over the iterations.

Typically ILC is used to feed the past errors along the trajectory (filtered and multiplied with a \emph{learning} matrix) back to the system for the next trial as feedforward compensations. A well designed feedback controller, whenever available, is only used to reject nonrepeating disturbances and to stabilize the system in the time domain. The recursive implementation \eqref{recursive_implementation}, on the other hand, readily provides and updates a feedback controller based on past performance. From here on, we will refer to the feedforward part of \eqref{recursive_implementation} as $\linInput_{k,j}$, keeping the feedback control separate.

\vspace{2mm}
\subsubsection{Computing the Expectations} The expectations appearing in $\eqref{recursive_ilc_cautious}$ can be calculated given the covariances $\var_{k,j}$ of the parameters,
\begin{equation}
\begin{aligned}
\matOne_{k,j} &= \tilde{\matOne}_{k,j} + \vec{R}_{j}, \\
\tilde{\matOne}_{k,j}^{a,b} &= \sum_{c=1}^{n}\sum_{d=1}^{n} \vec{P}_{k,j+1}^{c,d} \left(\mathbb{E}[\vec{B}_{k,j}^{c,a}]\mathbb{E}[\vec{B}_{k,j}^{d,b}] \!+\! \sigma(\vec{B}_{k,j}^{c,a},\vec{B}_{k,j}^{d,b})\right), \\
\matTwo_{k,j}^{a,b} &= \sum_{c=1}^{n}\sum_{d=1}^{n} \vec{P}_{k,j+1}^{c,d} \left(\mathbb{E}[\vec{B}_{k,j}^{c,a}]\mathbb{E}[\vec{A}_{k,j}^{d,b}] + \sigma(\vec{B}_{k,j}^{c,a},\vec{A}_{k,j}^{d,b})\right), \\
\matThree_{k,j}^{a,b} &= \sum_{c=1}^{n}\sum_{d=1}^{n} \vec{P}_{k,j+1}^{c,d} \left(\mathbb{E}[\vec{A}_{k,j}^{c,a}]\mathbb{E}[\vec{A}_{k,j}^{d,b}] + \sigma(\vec{A}_{k,j}^{c,a},\vec{A}_{k,j}^{d,b})\right),
\end{aligned}
\end{equation}
\noindent where the upper indices $a,b$ denote the corresponding entry of the matrix appearing on the left-hand side. The covariance matrices $\vec{\Sigma}_{k,j}$ contain the scalar covariance terms $\sigma(\cdot)$ on the relevant entries, i.e., 
\begin{equation}
\begin{aligned}
\sigma(\vec{A}_{k,j}^{c,a},\vec{A}_{k,j}^{d,b}) &= \left(\vec{\Sigma}_{k,j}\right)^{(a-1)n + c, (b-1)n + d}, \\
\sigma(\vec{B}_{k,j}^{c,a},\vec{A}_{k,j}^{d,b}) &= \left(\vec{\Sigma}_{k,j}\right)^{n^2 + (a-1)n + c, (b-1)n + d}, \\
\sigma(\vec{B}_{k,j}^{c,a},\vec{B}_{k,j}^{d,b}) &= \left(\vec{\Sigma}_{k,j}\right)^{n^2 + (a-1)n + c, n^2 + (b-1)n + d}. 
\end{aligned}
\end{equation}
\noindent The indexes of $\vec{B}_{k,j}$ covariances start from $n^2$ since the model matrix parameters in $\eqref{model-params}$ are vectorized starting from $\vec{A}_{k,j}$. 


\section{Online Implementation}\label{implementation}
In this section we algorithmically describe the recursive, adaptive and cautious $\alg$ proposed in the last two sections in detail, with the extensions for an online robot learning application. We will consider tracking table tennis trajectories as our application of choice. The online learning algorithm is readily applicable to similar dynamic tasks with real-time constraints, such as throwing, catching skills in sports or fast, demanding manufacturing tasks. 

The proposed ILC framework is summarized in Algorithm~\ref{alg_bayesILC}. Before entering the main loop (lines $7-16$), the trajectory is executed with inverse dynamics and time-varying LQR feedback (line $5$). The errors along the trajectory are filtered with a zero-phase filter (line $6$). During the cautious ILC update the feedback control law as well as the feedforward control inputs are updated recursively (line $9$). From the first iteration onwards, the means and the covariances of the model matrices are updated (line~$14$) before computing the feedforward input compensations $\linInput_{k,j}$ and the feedback matrices $\fbMat_{k,j}$. If the variant adaptation laws discussed in Section~\ref{adaptation} are employed, it will be enough to store the means and covariances of the relevant model parameters. These parameters can then be transformed, as discussed before, to form the discrete-time model matrix means and covariances, which are used in the cautious ILC update (line $9$).

Based on the forgetting factor $\forget$, the model adaptation strikes a balance between the prior model parameter distribution and the data observed in iteration~$k$. For the discrete LTV model and the link parameter adaptation, the data used is $\obs_{k,j} = \hat{\error}_{k,j+1} - \hat{\error}_{k-1,j+1}$. If continuous model matrix adaptation is performed, the data will instead be the smoothened joint acceleration differences, see \eqref{cts_model}. We discuss the effects of the forgetting factor and the different model adaptation strategies in more detail in Section~\ref{experiments}.

\begin{algorithm}[t!]
	\caption{Recursive, adaptive and cautious $\alg$.}
	\label{alg_bayesILC}
	\begin{algorithmic}[1]
		\Require $\dynamicsNominal$, $\traj_j$, $\forget, \epsilon > 0$, $\vec{Q}_j \succeq 0$, $\vec{R}_{j} \succ 0$, $\var_{0,j} \succ 0$
		\State Move to initial posture $\joint_0 = \traj_0$, $\dot{\joint}_0 = \vec{0}$.
		\State Initialize $k = 1$, $\linInput_{0,j} = \vec{0}$, $j = 1, \ldots, N$
		\State Compute mean dyn. parameters $\mean_{0,j}$ by linearizing $\dynamicsNominal$
		\State Compute feedback $\fbMat_{0,j} = \textrm{LQR}(\vec{Q}_j,\vec{R}_j,\mean_{0,j},\var_{0,j})$
		\State Execute with inv. dyn. $\trjInput$ and feedback $\fbMat_{0,j}$
		\State Filter errors with a zero-phase filter (output: $\hat{\error}_{0,j}$)
		\Repeat \Comment{\textcolor{gray}{ILC operation}}
		\State Compute error norm $\fullCost_k \!=\! \big(\sum_{j = 1}^{N} \! \hat{\error}^{\mathrm{T}}_{k,j}\vec{Q}_j\hat{\error}_{k,j}\big)^{\scriptscriptstyle{1/2}}$
		\State Compute $\linInput_{k,j}, \fbMat_{k,j}$ recursively using $\eqref{recursive_ilc_cautious}$ --	$\eqref{recursive_implementation}$
		\State Update feedforward controls $\sysInput_{k+1,j} = \sysInput_{k,j} + \linInput_{k,j}$
		\State Execute with $\sysInput_{\mathrm{IDM},j} + \sysInput_{k+1,j}$ and feedback $\fbMat_{k,j}$
		\State Observe errors $\error_{k,j} = \state_{k,j} - \traj_j$
		\State Filter errors with a zero-phase filter (output: $\hat{\error}_{k,j}$)
		\State Update model $\mean_{k,j}, \var_{k,j}$ using $\eqref{lbr_with_forgetting}$
		\State $k \gets k+1$
		\Until{$\fullCost_k < \epsilon$}
	\end{algorithmic}
\end{algorithm}


The practitioner, wary of the model inaccuracies, can increase robustness and ensure stability by setting large diagonal terms for the initial covariance of model uncertainty, $\var_{0,j} = \gamma\vec{I}, \, \gamma \gg 1, \, j \!=\! 1,\ldots,N$. Moreover, setting large covariances initially helps to observe the inaccuracies of the model and the noise statistics. The covariance will be suitably decreased over the iterations, as adaptation \eqref{lbr_with_forgetting} updates the linear models. Observing the noise statistics over the iterations can further help in the design of a good zero-phase filter to reject noise. Without accurate smoothing, adaptive ILC approaches run the risk of picking up noise in the adapted model matrices, which are then used in the control input update (in
our case, in equation \eqref{recursive_implementation}). This can hinder the control performance, hence we advice caution in the design of a smoothening filter.

The proposed update law takes advantage of the learning efficiency and computational advantages of model-based recursive ILC while being cautious with respect to model mismatch. The computational complexity of the recursive update is $\mathcal{O}(Nn^{3})$ as opposed to batch norm-optimal ILC, where the batch pseudoinverse operation typically incurs $\mathcal{O}(N^{3}n^{3})$ complexity. The batch model-based implementation using the \emph{lifted-vector form}~\cite{Bristow06} inverts the input-to-output matrix $\vec{F}$, 
\begin{equation}
\begin{aligned}
\vec{U}_{k+1} &= \vec{U}_{k} - \vec{F}^{\dagger}\vec{E}_k, \\
\vec{E}_k &= \left(\error_{k,1}^{\mathrm{T}}, \error_{k,2}^{\mathrm{T}}, \ldots, \error_{k,N}^{\mathrm{T}}\right)^{\mathrm{T}},
\end{aligned}
\label{batch-ilc}
\end{equation}
\noindent where the submatrices of the input-to-output matrix $\vec{F}$ are
\begin{equation}
\begin{aligned}
\vec{F}_{(i,j)} &= \left \{
\begin{array}{cc}
\vec{A}_{i-1}\ldots \vec{A}_j \vec{B}_{j-1}, & j < i, \\ 
\vec{B}_{j-1}, & j = i, \\
\vec{0}, & j > i. 
\end{array} \right.
\end{aligned}
\label{Fmatrix}
\end{equation}
The condition of the lifted model matrix \eqref{Fmatrix} grows exponentially with $N$ and inverting it quickly becomes numerically unstable.
\vspace{2mm}
\subsubsection{Implementation for Tracking Table Tennis Trajectories}\label{algorithm}

The online learning framework for robot table tennis is described in Algorithm~\ref{alg_tabletennis}. Whenever a ball is initialized from a fixed ballgun with constant settings, located at $\ball_0$, the trajectory generation framework will compute a particular striking trajectory (lines $2-3$) to intercept and hit the ball in real time. See Appendix ~\ref{hitting} for an overview of the trajectory generation pipeline. ILC can then be initialized (line $4$) by linearizing the dynamics model $\dynamicsNominal$ around the computed trajectory points $\traj_j, j = 1, \ldots,N$. ILC needs to be initialized only once, as long as the computed trajectory is capable of returning the ball to the opponent's court. The approximately $8 cm$ radius of the racket can cover for the inconstancy of the ballgun up to a certain degree. 

Whenever the striking trajectory is executed (line $6$), a returning trajectory will bring the arm back (line $7$) from the current state to the fixed initial posture, $\joint_0$. The returning trajectory can be as simple as a linear trajectory in the joint space. The \emph{consistency} provided by the fixed ballgun in our setup, shown in Figure~\ref{robot}, allows us to use ILC to track invariant trajectories over the iterations.

For a good performance in table tennis, the striking parts of these hitting movements need to be tracked accurately. The strikes are initially tracked with computed-torque inverse dynamics feedforward control commands and the additional LQR feedback. The feedback law is computed for this purpose by linearizing the nominal dynamics model around the striking part of the reference trajectory. After a strike is completed, feedback will switch to PD-gains for the returning trajectory and the arm will come back close to $\joint_0$. Learning with ILC can then take place (line 8) while waiting for another incoming table tennis ball.


%
\begin{algorithm}[t!]
\caption{ILC improving execution of robot table tennis hitting movements online.}
\label{alg_tabletennis}
\begin{algorithmic}[1]
	\Require $\joint_0$, $\dynamics_{\mathrm{ball}}$, $\alg(\ldots)$ (see Algorithm\ref{alg_bayesILC})
	\State Move to initial posture $\joint_0$, $\dot{\joint}_0 = \vec{0}$.
   	\State Predict ball trajectory $\ball_j$ using $\dynamics_{\mathrm{ball}}$
   	\State Compute trajectory $\traj_j$ given $\joint_{0}$ and $\ball_j$, $\scalebox{0.9}{j = 1, \ldots, N}$
 	\State Setup $\alg$ (lines $2-4$) 
	\Repeat \Comment{\textcolor{gray}{fixed ballgun throws balls at a constant rate}}
	   \State Execute strike with $\vec{u}_{\mathrm{ILC}}$ and LQR feedback $\fbMat$
	   \State Return to $\joint_0$ with high-gain PD control and linear traj.
	   \State Update $\vec{u}_{\mathrm{ILC}}$ and $\fbMat$ with $\alg$ (lines $9-14$)
   \Until{ballgun is moved}
	\end{algorithmic}
\end{algorithm}


The striking trajectory in table tennis is only an intermediary and does not need to be precisely tracked for a successful performance. In general, for hitting and catching tasks, the task performance depends critically on reaching the desired joint positions and velocities at the final time. A good performance along the trajectories is a means to this desired end: if feedback keeps the system stable around the trajectories, and the (linearized) models are reasonably valid around the trajectories, convergence to desirable performance levels can be rapid. 

%
\vspace{2mm}
\subsubsection{Coping with Varying Initial Conditions}\label{robustness_to_ic} Execution errors in tracking the reference trajectory (including the returning segment) prevents the robot from initializing in each iteration at the same state. Putting very high feedback gains on the returning trajectory or waiting long enough may suffice to initialize the system close to desired initial conditions, but in some occasions, none of these options may be desirable or available. For example, a robot \emph{practicing} table tennis with a fixed ballgun running at a fixed rate, may not have time to initialize its desired posture accurately. 

Starting from varying initial conditions $\state_{k,0} = [\joint^{\intercal}_{k,0},\dot{\joint}^{\intercal}_{k,0}]^{\mathrm{T}}$ one can consider updating the hitting movement $\traj_j$ to take the robot to the same hitting state. For such online updating of trajectories, the invariant trajectory parameters $\vec{p}$ can be used to generate the trajectory from the current joint values. The reference control inputs $\trjInput$ can then be recomputed based on the nominal inverse dynamics model. With this correction the total feedforward control commands $\sysInput_{\mathrm{ILC}}$ at iteration $k+1$ are re-computed as
\begin{equation}
\begin{aligned}
\sysInput_{\mathrm{ILC},j} &= \sysInput_{k+1,j} + \sysInput_{\mathrm{IDM},j}(\tilde{\traj}_j) - \sysInput_{\mathrm{IDM},j}(\traj_j), \\
\end{aligned}
\label{varying_ic}
\end{equation}
\noindent where $\tilde{\traj}_j$ is the updated trajectory starting from the perturbed initial state $\state_0 + \delta\state_{k,0}$.
Using this simple adjustment \eqref{varying_ic}, the stability of the learning performance can be greatly improved. 



\section{Evaluations and Experiments}\label{experiments}
%
In this section, we demonstrate the effectiveness of the ILC algorithm $\alg$ presented in Algorithm~\ref{alg_bayesILC} and described in detail in Section~\ref{implementation} in the context of tracking table tennis trajectories. We validate the proposed learning control law first in extensive simulations with linear and nonlinear models. In the last section we show real robot experiments with two seven degree of freedom Barrett WAM arms for tracking table tennis striking movements.
\subsection{Verification on Toy Problems}\label{simulations}
Stability is an important issue in the implementation of different learning controllers in real robot tasks. As a result, we setup extensive simulation experiments to validate the stability and robustness of our learning approach. We also discuss in detail the advantages of the recursive formulation over the batch pseudo-inverse ILC \eqref{batch-ilc}.
\vspace{2mm}
\subsubsection{Random Linear Models}
We generate here random linear models and random trajectories drawn from Gaussian Processes (GP) with squared exponential kernels. More specifically, the elements of the linear time-varying (LTV) model matrices $\vec{A}_j, \vec{B}_j$ are drawn from $(n+m)n$ uncorrelated GPs. The hyperparameters (scale, noise and smoothness parameters) of these GPs are drawn independently from normal distributions with fixed means and variances. Moreover, random perturbations of these models (drawn the same way from $(n+m)n$ uncorrelated GP's) are generated to construct nominal models. Using the proposed random disturbance generation scheme, we can average the results and construct error bars for different ILC algorithms.
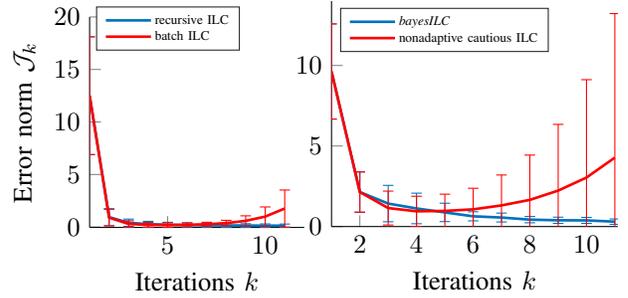
\begin{figure}[t!]
	\begin{minipage}{.2\textwidth}
	\setlength{\figureheight}{2.8cm}
	\setlength{\figurewidth}{3cm}
%
%
\definecolor{mycolor1}{rgb}{0.00000,0.44700,0.74100}%
\begin{tikzpicture}

\begin{axis}[%
width=0.951\figurewidth,
height=\figureheight,
at={(0\figurewidth,0\figureheight)},
scale only axis,
xmin=1,
xmax=12,
xlabel={Iterations $k$},
ymin=-0.0442217996701776,
ymax=20,
ylabel={Error norm $\fullCost_k$},
axis background/.style={fill=white},
axis x line*=bottom,
axis y line*=left,
legend style={at={(0.03,1.05)},anchor=north west,legend cell align=left,align=left,draw=white!15!black}
]
\addplot [color=mycolor1,solid,line width=1.0pt]
 plot [error bars/.cd, y dir = both, y explicit]
 table[row sep=crcr, y error plus index=2, y error minus index=3]{%
1	12.5021959324328	5.6064920934588	5.6064920934588\\
2	0.93757947671447	0.805217122827762	0.805217122827762\\
3	0.403582372029571	0.347827016540908	0.347827016540908\\
4	0.267792888533215	0.270534790217182	0.270534790217182\\
5	0.212467350652153	0.234068890760316	0.234068890760316\\
6	0.18283244844029	0.21004605732548	0.21004605732548\\
7	0.163481423067892	0.193289380011885	0.193289380011885\\
8	0.149544922971789	0.180766177593889	0.180766177593889\\
9	0.138857017200863	0.170907092760797	0.170907092760797\\
10	0.130296132196836	0.162846874479667	0.162846874479667\\
11	0.123213962594777	0.156078793775713	0.156078793775713\\
};
\addlegendentry{recursive ILC};

\addplot [color=red,solid,line width=1.0pt]
 plot [error bars/.cd, y dir = both, y explicit]
 table[row sep=crcr, y error plus index=2, y error minus index=3]{%
1	12.5021959324328	5.6064920934588	5.6064920934588\\
2	0.890230307012728	0.789945287319306	0.789945287319306\\
3	0.285277253383828	0.310503001433491	0.310503001433491\\
4	0.199193449226548	0.243415248896726	0.243415248896726\\
5	0.186939427107605	0.210460225527029	0.210460225527029\\
6	0.206977755358613	0.185502659751283	0.185502659751283\\
7	0.259877049338796	0.187843362907559	0.187843362907559\\
8	0.369919648722232	0.26500636135512	0.26500636135512\\
9	0.583446604761691	0.478549085268465	0.478549085268465\\
10	0.989487262964951	0.917779707618024	0.917779707618024\\
11	1.75901070926123	1.76230611809209	1.76230611809209\\
};
\addlegendentry{batch ILC};

\end{axis}
\end{tikzpicture}%
	\end{minipage} %
	\begin{minipage}{.2\textwidth}
	\setlength{\figureheight}{3cm}
	\setlength{\figurewidth}{4cm}
%
%
\definecolor{mycolor1}{rgb}{0.00000,0.44700,0.74100}%
\begin{tikzpicture}

\begin{axis}[%
width=0.951\figurewidth,
height=\figureheight,
at={(0\figurewidth,0\figureheight)},
scale only axis,
xmin=1,
xmax=11.1,
xlabel={Iterations $k$},
ymin=0,
ymax=14,
axis background/.style={fill=white},
axis x line*=bottom,
axis y line*=left,
legend style={at={(0.03,0.97)},anchor=north west,legend cell align=left,align=left,draw=white!15!black}
]
\addplot [color=mycolor1,solid,line width=1.0pt]
 plot [error bars/.cd, y dir = both, y explicit]
 table[row sep=crcr, y error plus index=2, y error minus index=3]{%
1	9.62732951118444	2.95417407341529	2.95417407341529\\
2	2.14292429416288	1.25047727533284	1.25047727533284\\
3	1.4273111327993	1.13047291804898	1.13047291804898\\
4	1.1244439814306	0.948930940273969	0.948930940273969\\
5	0.878351633910863	0.570428877890663	0.570428877890663\\
6	0.636341783403993	0.282175771774235	0.282175771774235\\
7	0.560496438766162	0.27560362398157	0.27560362398157\\
8	0.430292208194264	0.195793931297705	0.195793931297705\\
9	0.388959280686338	0.193750135145028	0.193750135145028\\
10	0.382666519638748	0.184978115234453	0.184978115234453\\
11	0.30402676446295	0.167047283146307	0.167047283146307\\
};
\addlegendentry{$\alg$};

\addplot [color=red,solid,line width=1.0pt]
 plot [error bars/.cd, y dir = both, y explicit]
 table[row sep=crcr, y error plus index=2, y error minus index=3]{%
1	9.62732951118444	2.95417407341529	2.95417407341529\\
2	2.14292429416288	1.25047727533284	1.25047727533284\\
3	1.14841798584659	1.05192959483815	1.05192959483815\\
4	0.946613275926693	0.935116405559931	0.935116405559931\\
5	0.971755511146965	1.03934820780272	1.03934820780272\\
6	1.06847255873286	1.30909634082327	1.30909634082327\\
7	1.31282781053405	1.8883043829723	1.8883043829723\\
8	1.65649792302105	2.78200875138299	2.78200875138299\\
9	2.23186063398009	4.11203134027657	4.11203134027657\\
10	3.0422178347523	6.07762998906434	6.07762998906434\\
11	4.28048211694095	8.94431690273964	8.94431690273964\\
};
\addlegendentry{nonadaptive cautious ILC};

\end{axis}
\end{tikzpicture}%
	\end{minipage}
	\caption{ILC in recursive form is evaluated on random linear time-varying (LTV) systems. The Frobenius norm of the trajectory deviations, $\fullCost_k$, is plotted over the iterations $k$. Results are averaged over ten experiments, where for each experiment, trajectories, nominal models and actual models are drawn randomly from Gaussian Processes. The performance of the batch pseudo-inverse ILC \eqref{batch-ilc} is shown in the red line. Numerical stability issues prevent it from stabilizing at steady state error, whereas recursive ILC (blue line) converges stably. If the model mismatch is increased, at some point, recursive ILC also diverges. Applying caution without adaptation is not enough to converge to steady state error. Cautious \emph{and} adaptive $\alg$, on the other hand, applying the updates \eqref{update_model_matrices} and \eqref{recursive_implementation} iteratively, is very effective and shows a stably convergent behaviour.}
	\label{rand_lin_conv}
\end{figure}

The performance of the recursive implementation (i.e., Equation~\eqref{recursive_implementation} with zero covariances and no adaptation) is shown in Figure~\ref{rand_lin_conv} on the left-hand side, where the results are averaged over ten different trajectories and models. The dimensions of the models are $n = 2, m = 2$, and the horizon size is set to $N = 120$. For the LQR and ILC calculations, $\vec{R} = 10^{-6}\vec{I}$ and the weighting matrix $\vec{Q}$ was set to the identity. In this case, the batch model-based implementation using the pseudo-inverse \eqref{batch-ilc} is not stable at all without feedback. Applying LQR feedback and adding current iteration ILC in Figure~\ref{rand_lin_conv} improves the performance (red line in Figure~\ref{rand_lin_conv}), but numerical issues (i.e., large condition number) in inverting the large model matrix $\vec{F}$ in lifted form \eqref{Fmatrix} prevents it from stabilizing at steady state error. Tracking performance throughout the experiments is measured with respect to the Frobenius norm of the deviations $\vec{e}_{k,j}$, denoted as $\fullCost_{k}$.

For the simulation results in Figure~\ref{rand_lin_conv}, the spectral norm of the difference between the nominal and the actual models are each set to $\alpha \sigma_{\textrm{min}}(\vec{F})$ where $\alpha = 100$. Increasing $\alpha$ further increases the probability that the model-based ILC is not monotonically convergent for some trial. For example, one can observe \emph{asympotically} but not \emph{monotonically} convergent ILC behaviour when setting $\alpha = 990$ for a particular model and trajectory shown in Figure~\ref{lin_mc_sim_example}. Increasing $\alpha$ futher can prevent even asymptotic stability. 

Especially in these cases of high model mismatch, the proposed adaptive and cautious $\alg$ offers a stable and convergent ILC behaviour. In Figure~\ref{rand_lin_conv} on the right-hand side, we consider the case where $\alpha = 1000$. Recursive ILC that is also cautious does not show a stable convergent behaviour, whereas recursive ILC that is not cautious (i.e., covariance of the LTV matrices are zero) is not stable at all. Cautious \emph{and} adaptive $\alg$, on the other hand, using LBR ($\forget = 1.0$) to update the discrete-time LTV matrices $\vec{A}_{k,j}, \vec{B}_{k,j}$, shows a monotonic learning performance. The results are again averaged over ten different models and ten trajectories. For LBR, the initial covariances in \eqref{update_model_matrices} are set to $\vec{\Sigma}_{0,j} = \gamma\vec{I}$ for all $j = 1, \ldots, N$, where $\gamma = 10^{4}$ and the noise covariance is $\sigma^{2} = 1$. Changing the exponent of the initial covariance, or reducing the forgetting factor $\lambda$ in this case, can lead to a reduced or unstable learning performance.
\vspace{2mm}
\subsubsection{Gaussian Process Dynamics}
The performance of the proposed algorithm $\alg$ is evaluated next over random nonlinear models. In these set of experiments, we sample the states from $n$ uncorrelated GPs with squared exponential kernels and random linear mean functions. The hyperparameters of these GPs are randomized as before. By sampling from such random nonlinear models, we can test the proposed algorithm under nonlinear uncertainties and noisy outputs. The \emph{actual} model is simulated as follows:
\begin{enumerate}
	\item Random reference control inputs $\vec{v}_j \in \mathbb{R}^{m}, j = 1, \ldots, N$ are drawn $K$ times from $m$ \emph{control} GPs.
	\item $n$ \emph{oracle} GPs are used to sample $\dynamics(\state_j, \vec{v}_j)$ and the generated dynamics is integrated (starting from zero initial conditions) using forward Euler, $dt = 1/N$, to form $K$ trajectories. The GPs are conditioned during this process on the generated states $\state_j$ and inputs $\vec{v}_j$.
\end{enumerate}
These $n$ oracle GPs constitute the \emph{actual} but unknown nonlinear dynamics model. Nominal models can be easily generated by using the predictions of the oracle GPs at a subset of the state space. The construction of a \emph{nominal} model is described in detail below:
\begin{enumerate}
	\item Another set of control inputs $\sysInput_j, \ j = 1, \ldots, N$ are drawn from the \emph{control} GPs, as before.
	\item The mean predictions $\dynamics(\state_j, \sysInput_j)$ of the oracle GPs at $\sysInput_j$ are used to evolve these control inputs (as in step 2 of the actual model).
	\item The $n$ separate \emph{model} GPs (with same hyperparameters as the oracle) are conditioned on the resulting trajectory, i.e., the input pairs $(\state_j, \sysInput_j)$ and the outputs $\dynamics_j = (\state_{j+1} - \state_{j}) / dt$ for each time step $j = 1, \ldots, N$.
	\item The mean derivative of the model GPs are calculated analytically (using the kernel derivatives). Discretized time-varying matrices $\vec{A}_j, \vec{B}_j$ and their variances $\vec{\Sigma}_{0,j}$ are constructed for each $j = 1, \ldots, N$, based on the mean and variance of the GP derivatives.
\end{enumerate}
By sampling $K = 20$ trajectories for the conditioning of oracle GPs, we can cover a significant part of the state space in $n=2$ dimensions. For each ILC iteration thereafter, the mean predictions are used as in step $(2)$ to evolve the trajectory, but without further conditioning of the model GPs. Instead, adaptation is performed as before with LBR, replacing the steps $(3-4)$. We can thus avoid the expensive online GP training.

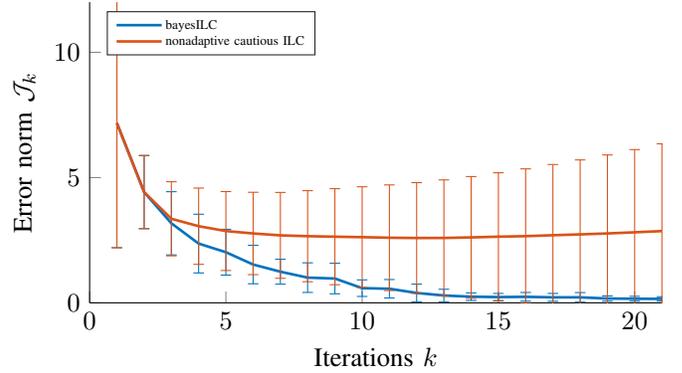
\begin{figure}[t!]
		\setlength{\figureheight}{4cm}
		\setlength{\figurewidth}{8cm}
%
%
\definecolor{mycolor1}{rgb}{0.00000,0.44700,0.74100}%
\definecolor{mycolor2}{rgb}{0.85098,0.32549,0.09804}%
\begin{tikzpicture}

\begin{axis}[%
width=0.951\figurewidth,
height=\figureheight,
at={(0\figurewidth,0\figureheight)},
scale only axis,
xmin=0,
xmax=21,
xlabel={Iterations $k$},
ymin=0,
ymax=12,
ylabel={Error norm $\fullCost_k$},
axis background/.style={fill=white},
axis x line*=bottom,
axis y line*=left,
legend style={at={(0.03,0.97)},anchor=north west,legend cell align=left,align=left,draw=white!15!black}
]
\addplot [color=mycolor1,solid,line width=1.0pt]
 plot [error bars/.cd, y dir = both, y explicit]
 table[row sep=crcr, y error plus index=2, y error minus index=3]{%
1	7.17550415607417	4.97577811593219	4.97577811593219\\
2	4.41997050876555	1.46072723806402	1.46072723806402\\
3	3.17452918727501	1.26153587141107	1.26153587141107\\
4	2.36553131739918	1.17187401385859	1.17187401385859\\
5	2.02108514619861	0.915640444923008	0.915640444923008\\
6	1.52882392026278	0.767677471624311	0.767677471624311\\
7	1.24344486889978	0.498953818469082	0.498953818469082\\
8	1.00624585381185	0.591155945465867	0.591155945465867\\
9	0.970628536784621	0.609857307861332	0.609857307861332\\
10	0.586031226802987	0.330810791521726	0.330810791521726\\
11	0.56268174512209	0.367958130186116	0.367958130186116\\
12	0.394822093612826	0.352343708659105	0.352343708659105\\
13	0.292047274477596	0.250239934967932	0.250239934967932\\
14	0.245549046664561	0.151611542531444	0.151611542531444\\
15	0.229455058895968	0.145175061183837	0.145175061183837\\
16	0.239922909132177	0.174734927343296	0.174734927343296\\
17	0.219170723313063	0.156607802608023	0.156607802608023\\
18	0.219464001332401	0.190053626636284	0.190053626636284\\
19	0.172478045494018	0.105341589017596	0.105341589017596\\
20	0.163119489088746	0.097477424573156	0.097477424573156\\
21	0.157452393112694	0.087866714067555	0.087866714067555\\
};
\addlegendentry{bayesILC};

\addplot [color=mycolor2,solid,line width=1.0pt]
 plot [error bars/.cd, y dir = both, y explicit]
 table[row sep=crcr, y error plus index=2, y error minus index=3]{%
1	7.17550415607417	4.97577811593219	4.97577811593219\\
2	4.41997050876555	1.46072723806402	1.46072723806402\\
3	3.35487071014708	1.47960140356536	1.47960140356536\\
4	3.05910098863031	1.52208611452889	1.52208611452889\\
5	2.86579691485635	1.57524060479907	1.57524060479907\\
6	2.76956126405292	1.64338603112441	1.64338603112441\\
7	2.69311619226394	1.71386318881998	1.71386318881998\\
8	2.66008062288291	1.81698276158383	1.81698276158383\\
9	2.63968805375169	1.91639143913412	1.91639143913412\\
10	2.62180743606883	2.01019242943187	2.01019242943187\\
11	2.59872738009565	2.10829828471085	2.10829828471085\\
12	2.58768722470095	2.21042907201435	2.21042907201435\\
13	2.59001279110657	2.31703011410875	2.31703011410875\\
14	2.61207904495434	2.42857853333372	2.42857853333372\\
15	2.64111704935795	2.55086287938626	2.55086287938626\\
16	2.66122500275163	2.68593259124026	2.68593259124026\\
17	2.69868728967652	2.8231849091916	2.8231849091916\\
18	2.73215896807878	2.9721360344803	2.9721360344803\\
19	2.7697067741936	3.13266556222154	3.13266556222154\\
20	2.81524533726158	3.30229650455801	3.30229650455801\\
21	2.86384578487144	3.48313689019474	3.48313689019474\\
};
\addlegendentry{nonadaptive cautious ILC};

\end{axis}
\end{tikzpicture}%
		\label{random_nonlinear_conv}
	\caption{The proposed ILC algorithm is evaluated on random nonlinear systems. The Frobenius norm of the trajectory deviations, $\fullCost_k$, is plotted over the iterations $k$. Results are averaged over ten experiments, where for each experiment, trajectories and dynamics along these trajectories are drawn from Gaussian Processes. Recursive ILC that is not cautious shows an unstable behaviour, and adding adaptation without caution is also not stable (both not shown in the Figure). Purely cautious ILC (red line) is divergent for some of the trajectories. Cautious \emph{and} adaptive $\alg$, on the other hand (blue line), shows a stable convergent learning performance. }
	\label{rand_nonlin_conv}
\end{figure}

Figure~\ref{rand_nonlin_conv} shows the learning performance for a horizon size of $N = 20$. The dimensions of the system is same as before, $n = 2, m = 2$ and $\vec{R} = 10^{-6}\vec{I}$, $\vec{Q} = \vec{I}$. The results are averaged again over ten experiments. 
%
In this nonlinear setting, the recursive ILC that is not cautious shows an unstable behaviour (not shown in Figure~\ref{rand_nonlin_conv}). Adaptive but not cautious ILC is also unstable (also not shown). Cautious but not adaptive ILC is not stable for some trajectories and can diverge (red line). Cautious \emph{and} adaptive $\alg$, on the other hand (blue line), using LBR to update the discrete-time LTV matrices, shows again a stable convergent learning performance, improving over the purely cautious ILC. For LBR, the initial covariances in \eqref{update_model_matrices} are again set to $\gamma = 10^{4}$ times the identity and the noise covariance is $\sigma^{2} = 1$. The best performance is reached when the forgetting factor is set to $\forget = 0.9$. As before, changing the exponent of the initial covariance, or the forgetting factor, can lead to a reduced or unstable learning performance.
\vspace{2mm}
\subsubsection{Barrett WAM Model}

We next test ILC on striking movements~\eqref{strike_traj} for a seven degree of freedom Barrett WAM simulation model. In the simulations, the robot is started from a fixed initial state $\joint_0$. The initial posture is chosen from one of the center, right-hand side or left-hand side resting postures of the robot. The striking parameters~\eqref{coeffs} are then optimized, based on an incoming table tennis ball with a randomly chosen incoming position and velocity. The link parameters of the Barrett WAM forward dynamics model used to simulate actual trajectories are perturbed randomly to construct nominal models for ILC. The linearization procedure described in Section~\ref{problem_statement} produces LTV nominal models that can be used by ILC to reduce the deviations from the desired (fixed) striking movement over the iterations.

\begin{figure}[t!]
	\begin{minipage}{.2\textwidth}
		\setlength{\figureheight}{3cm}
		\setlength{\figurewidth}{4cm}
%
%
\definecolor{mycolor1}{rgb}{0.00000,0.44700,0.74100}%
\definecolor{mycolor2}{rgb}{0.85000,0.32500,0.09800}%
\begin{tikzpicture}

\begin{axis}[%
width=0.951\figurewidth,
height=\figureheight,
at={(0\figurewidth,0\figureheight)},
scale only axis,
xmin=0.5,
xmax=11.5,
ymin=0,
ymax=2.5,
xlabel={Iterations $k$},
ylabel={Error norm $\fullCost_k$},
axis background/.style={fill=white},
axis x line*=bottom,
axis y line*=left,
legend style={nodes={scale=0.8, transform shape}, at={(1.05,0.95)}, anchor=north east, legend cell align=left, align=left, draw=white!15!black}
]
\addplot [color=mycolor1, line width=1.0pt]
 plot [error bars/.cd, y dir = both, y explicit]
 table[row sep=crcr, y error plus index=2, y error minus index=3]{%
1	1.5504419466451	0.269340822692037	0.269340822692037\\
2	1.48907114681406	0.212574015800399	0.212574015800399\\
3	1.13337105847224	0.209574195997255	0.209574195997255\\
4	1.07746851525526	0.619805250252477	0.619805250252477\\
5	1.21402639603789	0.476333305882405	0.476333305882405\\
6	0.732009076414156	0.294826384905428	0.294826384905428\\
7	0.752573742669798	0.377440593583141	0.377440593583141\\
8	0.572827714857623	0.160021209799356	0.160021209799356\\
9	0.560379489740966	0.185232077559164	0.185232077559164\\
10	0.467467277118401	0.121455916934696	0.121455916934696\\
11	0.372472886269765	0.0538436177144902	0.0538436177144902\\
};
\addlegendentry{Discrete LTV}

\addplot [color=mycolor2, line width=1.0pt]
 plot [error bars/.cd, y dir = both, y explicit]
 table[row sep=crcr, y error plus index=2, y error minus index=3]{%
1	1.5504419466451	0.269340822692037	0.269340822692037\\
2	1.48907114681406	0.212574015800399	0.212574015800399\\
3	1.66352629192718	0.493478710735085	0.493478710735085\\
4	1.24650762524959	0.605002000227847	0.605002000227847\\
5	1.02143526229188	0.553747608087374	0.553747608087374\\
6	1.228497296631	0.632045091709894	0.632045091709894\\
7	0.778090142890366	0.341130339875735	0.341130339875735\\
8	0.836784463765598	0.434274662261059	0.434274662261059\\
9	0.649709932328481	0.254967701142407	0.254967701142407\\
10	0.430330770805789	0.0913581172903587	0.0913581172903587\\
11	0.3641384742389	0.0554973406457824	0.0554973406457824\\
};
\addlegendentry{Continuous LTV}

\addplot [color=black, line width=1.0pt]
 plot [error bars/.cd, y dir = both, y explicit]
 table[row sep=crcr, y error plus index=2, y error minus index=3]{%
1	1.5504419466451	0.269340822692037	0.269340822692037\\
2	1.48907114681406	0.212574015800399	0.212574015800399\\
3	1.54906374183231	0.656517599209378	0.656517599209378\\
4	1.31148255627459	0.49076974244694	0.49076974244694\\
5	0.777205099057496	0.313187477278336	0.313187477278336\\
6	0.602107962684816	0.197149133268519	0.197149133268519\\
7	0.396020001250051	0.0916768611451872	0.0916768611451872\\
8	0.343522305037041	0.0756041628755282	0.0756041628755282\\
9	0.309062379087267	0.0796825399680765	0.0796825399680765\\
10	0.293397900254924	0.0790298174192346	0.0790298174192346\\
11	0.28580396176538	0.0779686572685741	0.0779686572685741\\
};
\addlegendentry{Link parameters}

\end{axis}
\end{tikzpicture}%
	\end{minipage}\qquad
	\begin{minipage}{.2\textwidth}
		\centering
		\setlength{\figureheight}{3cm}
		\setlength{\figurewidth}{3cm}
		\input{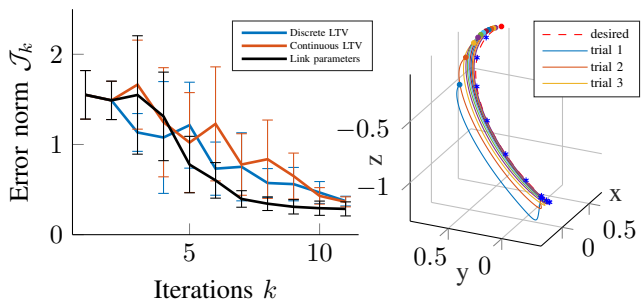}
	\end{minipage}
	\caption{The performance of the adaptive and cautious ILC algorithm $\alg$ on the simulated Barrett WAM model is shown on the left-hand side. The Frobenius norm of the trajectory deviations, $\fullCost_k$, is plotted over the iterations $k$. The results are averaged over ten different strikes and three different initial postures. Three different adaptation laws are considered, adaptation of discrete-time and continuous-time LTV models are shown in blue and red, respectively, while the adaptation of link parameters is shown in black. Forgetting factor was set to $\forget = 0.8$ for all of the adaptation laws. One of the desired trajectories, shown in dashed red on the right-hand side, is tracked very closely in the final iteration. The blue markers correspond to the time profile of the motion, which are drawn uniformly spaced, one for each $80$ milliseconds. The final hitting positions reached are shown as filled circles.}
	\label{wam_conv}
\end{figure}

The randomization during the optimization guarantees that a variety of hitting movements are tracked throughout the experiments. The performance of the proposed ILC approach $\alg$ with three different adaptation laws is then evaluated over the striking segment of the optimized (striking and returning) trajectories. The convergence results are averaged over ten such striking movements, as shown in Figure~\ref{wam_conv}. The adaptation of discrete-time and continuous-time LTV models are shown in blue and red, respectively, while the adaptation of link parameters is shown in black. Forgetting factor was set to $\forget = 0.8$ for all of the adaptation laws. Initial covariances are set to $\var_{0,j} = 10^4\vec{I}$ for continuous and discrete-time LTV model adaptation laws, while for link parameters, the initial covariances are $\var_{0,j} = 10^{10}\vec{I}$. The weights of the cautious ILC update \eqref{recursive_implementation} is set to $\vec{R} = 10^{-2}\vec{I}$, $\vec{Q} = \vec{I}$.

After updating the link parameter means and variances, we use an auto-differentiation tool (ADOL-C library in $\CC$) together with sampling to approximate the distribution of forward dynamics \eqref{dynamics} derivatives $\vec{A}_{k,j}, \vec{B}_{k,j}$. More specifically, the forward dynamics is differentiated (with respect to joint positions, velocities and control inputs) at $100$ link parameter samples drawn from the posterior distribution (i.e., normal distribution with means and variances given by \eqref{update_model_matrices}) online. This sampling procedure generates a reasonable approximation of posterior derivative means and variances.
 
\begin{figure}[t!]
	\scriptsize
	\centering 
	\setlength{\figureheight}{6cm}
	\setlength{\figurewidth}{7.5cm}
	\input{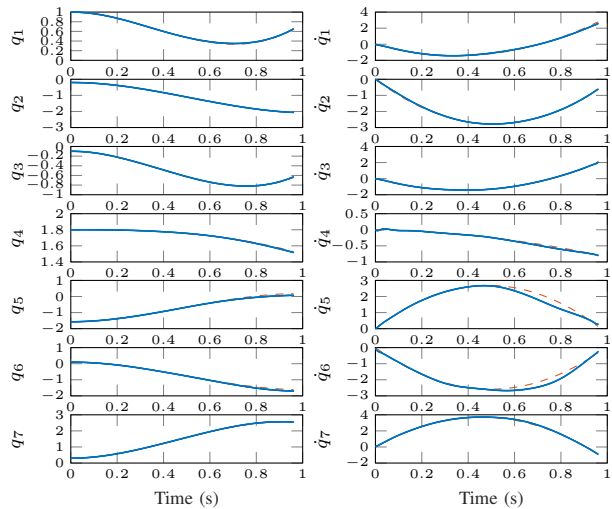}
	\caption{Joint trajectories for a hitting movement on the Barrett WAM model. The reference trajectories, shown in dashed red, are tracked very closely with ILC in the final iteration, shown in blue.}
	\label{wam_joint_traj_sim}
	\normalsize
\end{figure}

In table tennis, if the robot arm follows the assigned reference trajectory precisely it will hit the ball with a desired velocity at the desired time. We can see on the right-hand side of Figure~\ref{wam_conv} that an initial attempt (blue curve) falls short of the reference trajectories (dashed curve). The percentage of the balls that are returned to the opponent's court are close to zero. ILC then modifies the control inputs to compensate for the modeling errors. In the last attempt the reference trajectory is executed almost perfectly. The accuracy of the table tennis task increases to $\% 95$, on average. Figure~\ref{wam_joint_traj_sim} shows the adjusted control inputs for one striking movement. 

The recursive ILC (without adaptation or caution) is convergent for some of the hitting movements in Figure~\ref{wam_conv}. However, similar to the previous simulation examples, the recursive form of the ILC update, depending on the accuracy of the model along the trajectories, can fail to converge for some trajectories (not shown in the Figure). The proposed recursive, adaptive and cautious algorithm $\alg$, with the three adaptation laws shown in Figure~\ref{wam_conv}, shows a better and faster convergence, for a variety of trajectories. 

The ILC experiments shown in Figures~\mbox{\ref{wam_conv}--\ref{wam_joint_traj_sim}} reset the initial posture always to the same desired posture $\joint_0$. Next, we consider non-repetitive disturbances around the desired initial posture. This would mean, physically, that the robot is not initialized accurately around the resting posture.

\begin{figure}
	\center
	\setlength\figurewidth{4cm}
	\setlength\figureheight{4cm}
%
%
\definecolor{mycolor1}{rgb}{0.00000,0.44700,0.74100}%
\definecolor{mycolor2}{rgb}{0.85000,0.32500,0.09800}%
\begin{tikzpicture}

\begin{axis}[%
width=0.951\figurewidth,
height=\figureheight,
at={(0\figurewidth,0\figureheight)},
scale only axis,
xmin=1,
xmax=5,
xtick={1, 2, 3, 4, 5},
xlabel={Iterations},
ymin=0,
ymax=3.5,
ylabel={Final cost},
axis background/.style={fill=white},
axis x line*=bottom,
axis y line*=left,
legend style={legend cell align=left,align=left,draw=white!15!black,at={(0.05,0.9)}}
]
\addplot [color=mycolor1,solid,line width=1.0pt,mark=asterisk,mark options={solid}]
 plot [error bars/.cd, y dir = both, y explicit]
 table[row sep=crcr, y error plus index=2, y error minus index=3]{%
1	1.82641122021916	0.191978860965294	0.191978860965294\\
2	0.261054172108418	0.0249404822302659	0.0249404822302659\\
3	0.354990962550494	0.0838422312772416	0.0838422312772416\\
4	0.371762935698725	0.154244787869398	0.154244787869398\\
5	0.0645006335245202	0.047776901839287	0.047776901839287\\
};
\addlegendentry{ILC with trajectory adaptation};

\addplot [color=black,solid,line width=1.0pt,mark=diamond,mark options={solid}]
 plot [error bars/.cd, y dir = both, y explicit]
 table[row sep=crcr, y error plus index=2, y error minus index=3]{%
1	2.70121798595486	0.51084050072295	0.51084050072295\\
2	0.935515994089551	0.755322525630824	0.755322525630824\\
3	1.01957603259838	0.283347434829148	0.283347434829148\\
4	1.03639294861837	0.42746235529782	0.42746235529782\\
5	1.69140430251211	1.59514958515428	1.59514958515428\\
};
\addlegendentry{ILC without trajectory adaptation};

\end{axis}
\end{tikzpicture}%
	\caption{Simulation results illustrating the additional robustness to varying initial conditions whenever the trajectories (states and control references) are adapted according to \eqref{varying_ic} (blue line). Note the unstable performance of ILC without such adaptation (black line), which keeps the references $\traj_j$ and the inverse dynamics inputs $\sysInput_{\mathrm{IDM},j}$ fixed.}
	\label{wam_goal_based_ilc_sim}
\end{figure}
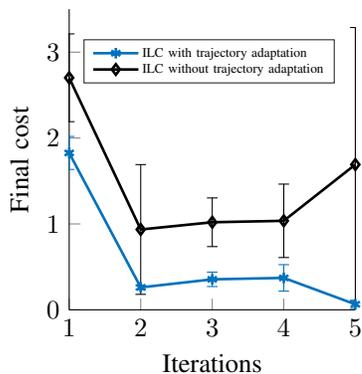

Comparisons to the baseline (black line) in Figure~\ref{wam_goal_based_ilc_sim} illustrate the additional robustness whenever the trajectory adaptation \eqref{varying_ic} is employed. We adapt the metric for this comparison according to the task: the costs indicated are the \emph{final} costs (for hitting the incoming ball at the desired joint positions with desired joint velocities), not the full costs incurred along the reference trajectory. Note especially the faster convergence and increased accuracy of the proposed method with the reference trajectory and input adaptation (blue line). More robust performance is obtained by adapting the trajectories $\traj_j$ and $\sysInput_{\mathrm{IDM},j}$, which, in addition to performing better, shows much lower variance compared to the baseline.

In practice trusting the model too much at the beginning of the trajectory leads to the amplification of initial errors. Nonrepetitive starting postures violate the initial condition assumption typical of standard ILC updates. In this case, the feedback matrices $\fbMat_{k,j}$, as opposed to the feedforward input updates $\linInput_{k,j}$, play a bigger role in the learning stability at the beginning of the trajectories, $j \ll N$.
%

\subsection{Real Robot Table Tennis}
%

\begin{figure*}
	\centering
	\includegraphics[height=9cm,width=16cm,scale=0.25]{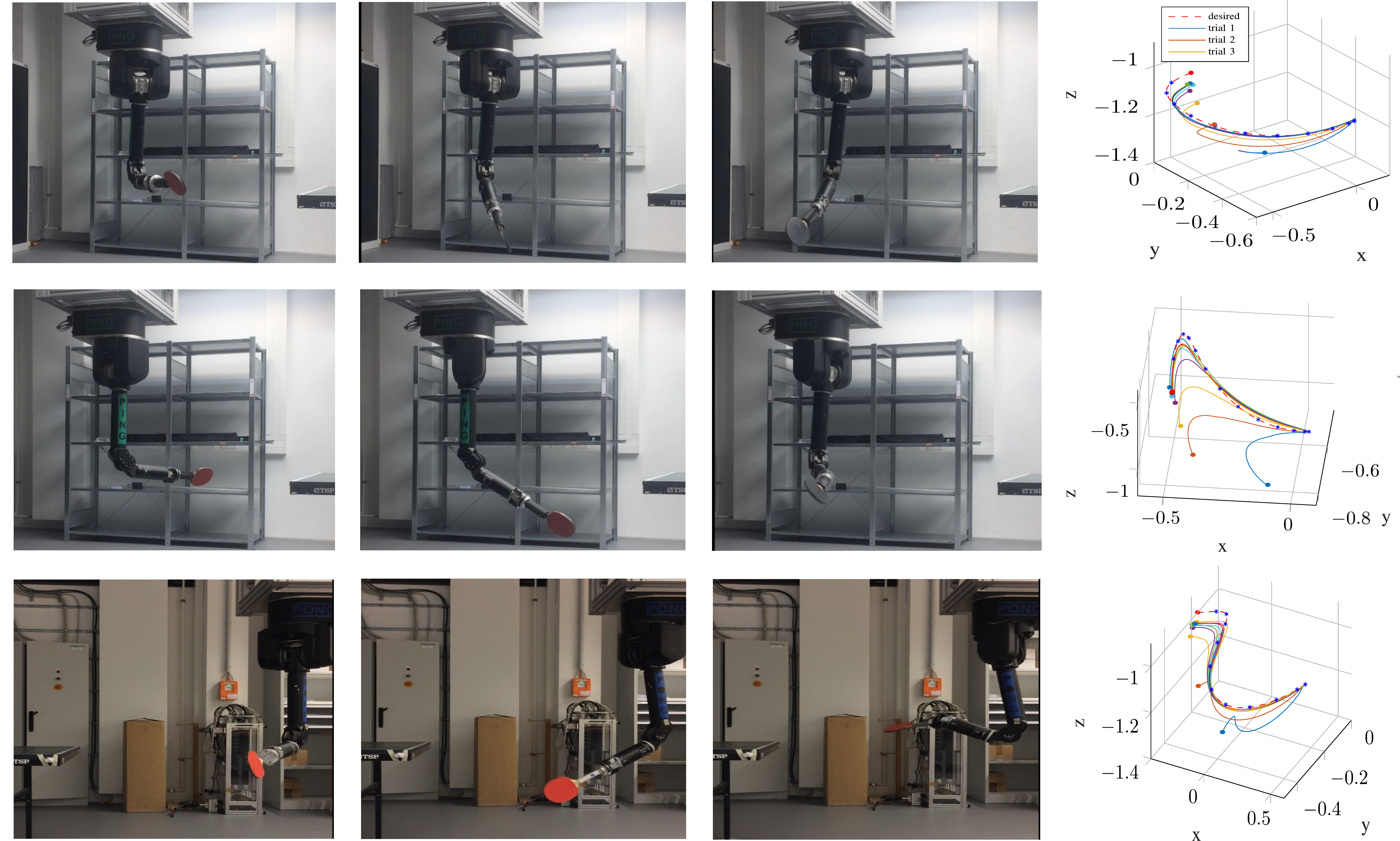}
	\caption{Two Barrett WAMs (a.k.a. \emph{Ping} and \emph{Pong}) are initialized in our experiments in three different starting postures. We make controlled experiments with a simulated ballgun, and generate many different hitting movements, three of them are shown in the above images. The proposed algorithm $\alg$ leads to an efficient and stable learning approach for tracking these hitting movements. The right-hand side starting posture for the robot Ping can be seen on the upper left image. Initially, before learning with ILC starts, Ping performs poorly, and the hitting posture of the robot is shown in the upper central image. After five iterations, the hitting posture is corrected significantly as shown in the upper right image. Similarly, the central images show the operation of the ILC for another trajectory, where the starting posture for Ping is fixed on the left-hand side of the robot. On the bottom images, an ILC performance is shown for the robot Pong. The three plots on the right-hand side show the Cartesian trajectories corresponding to the ILC iterations. The reference trajectories are shown in dashed red, and the final hitting positions reached are shown as filled circles.}
	\label{ping_pong_moving}
\end{figure*}

Finally we perform experiments on our robotic table tennis platform, see Figure~\ref{ping_pong_moving}, where two seven degree of freedom (DoF) cable-driven, torque-controlled Barrett WAM arms (\emph{Ping} and \emph{Pong}) are hanging from the ceiling. The custom made Barrett WAM arms are capable of high speeds and accelerations (approx. up to $10 m/s^2$ in task space). Standard size rackets (16 cm diameter) are mounted on the end-effector of the arms as can be seen in Figure~\ref{ping_pong_moving}. A vision system consisting of four cameras hanging from the ceiling around each corner of the table is used for tracking the ball \cite{Lampert12}. A ball launcher (see Figure~\ref{robot}) is available to throw balls accurately to a fixed position inside the workspace of the robots. The incoming ball arrives with low-variability in desired positions and higher-variability in ball velocities. The whole area to be covered amounts to about 1 m$^2$ circular region surrounding an initial centered posture of the robots.

%
\begin{figure}[b!]
	\setlength{\figureheight}{4cm}
	\setlength{\figurewidth}{6cm}
	\input{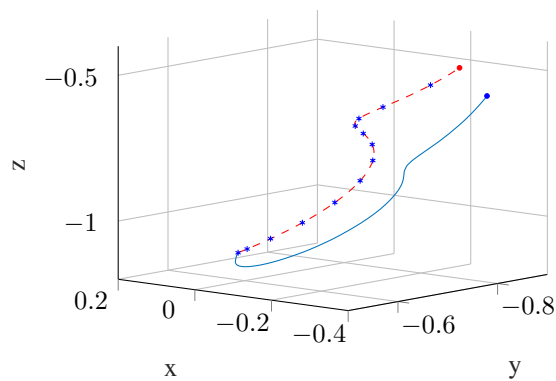}
	\caption{An example of a striking movement for real robot table tennis is shown in red. The blue markers correspond to the time profile of the motion, which are drawn uniformly, one for each $80$ milliseconds. Executing this movement well with the Barrett WAM will lead to a good hit. Control errors in tracking lead to a poor hitting performance, shown in blue. The filled circles are the final reached hitting positions. High-gain PD feedback was used to track the reference in this real robot example. The tracking errors can be decreased efficiently and stably by applying the proposed recursive, cautious and adaptive ILC update $\alg$.} 
	\label{tracking-with-pd}
\end{figure}
The realistic simulation environment SL~\cite{Schaal06} acts as both a simulator and as a real-time interface to the Barrett WAMs in our experiments. The initial positioning is given by a PD controller with high gains on the shoulder joints, which is then toggled off during the experiments with the striking movements, as summarized in Algorithm~\ref{alg_tabletennis}. The high-gain PD controller used to initialize the robots was also tested for tracking the striking movements, see Figure~\ref{tracking-with-pd}. When ILC is applied on top of the PD controller, the learning quickly stagnates, leading to oscillations in some of the joints. Instead, a low-gain LQR feedback law is computed for the striking part of the movement with a linearized nominal dynamics model \eqref{discreteLTV}. The weighting matrices for this purpose are set to identity, $\vec{Q} = \vec{I}$, and the constant penalty matrix is chosen as $\vec{R} = 0.05\vec{I}$. Decreasing the scaling of the penalty matrix to $0.03$ causes oscillations in the elbow joint, indicating that the nominal model is not very accurate. At the cost of larger initial error, we suggest increasing the input penalties $\vec{R}$ to improve the stability of ILC in other high degrees-of-freedom robotics applications.


After the visual system outputs a ball estimate, a ball model can be used along with an Extended Kalman Filter to predict a ball trajectory. The ball model accounts for some of the bouncing behavior of the ball and air drag effects. If the predicted ball trajectory coincides with the workspace of the robot, the motion planning system has to then come up with a trajectory that specifies how, where and when to intercept the incoming ball. Desired Cartesian position, velocity and orientations of the racket at the hitting time $T$ impose constraints on the seven joint angles and seven joint velocities of the robot arm at $T$. Along with the desired hitting time $T$ (or the time until impact), these fifteen parameters are used to generate third-order joint space polynomials. These movements can be optimized online in $20-30$ milliseconds~\cite{Koc18}, or loaded from a lookup table. In the ILC experiments, the parameters in the lookup table are used without interpolation, to make sure that the same trajectory can be used for balls deviating slightly from their stored values. 
We make sure that the lookup table is dense enough and that the ballgun is fixed.

%


%
\begin{figure}[t!]
	\centering 
	\setlength{\figureheight}{10cm}
	\setlength{\figurewidth}{7.5cm}
	\input{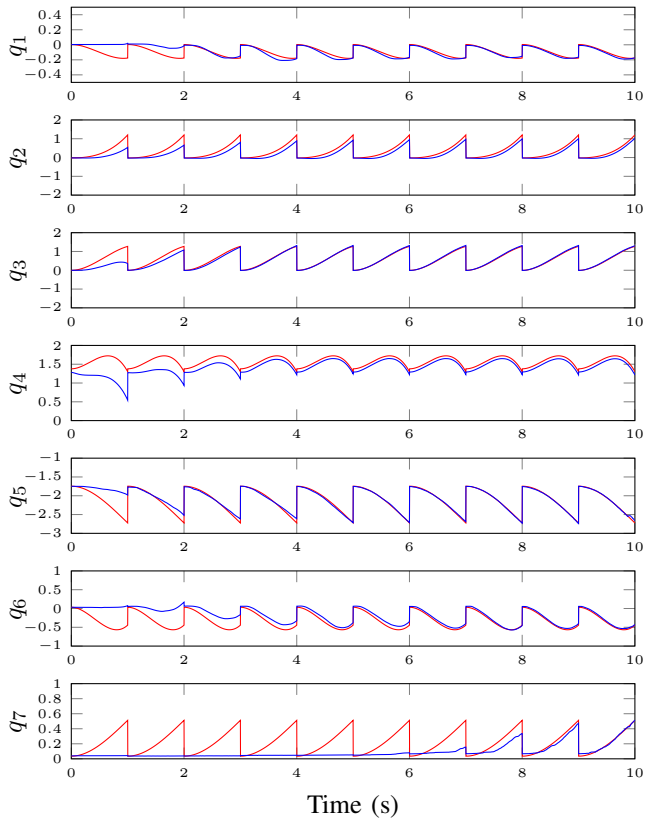}
	\caption{Robot experiment results for cautious and adaptive $\alg$, shown for a particular reference trajectory. The ten iteration results are concatenated for convenience. The desired joint trajectories correspond to a hitting movement on the Barrett WAM. The reference trajectories, shown in red, are tracked very closely with ILC in the final iteration, shown in blue. Final cost goes down to 0.20 in the last iteration.}
	\label{real_wam_joint_traj}
\end{figure}
%

Some examples of the generated trajectories are shown in Figure~\ref{ping_pong_moving}. After a strike, a linear joint trajectory is computed that will take the robots from the current state to the resting posture in $T_{\mathrm{rest}} = 1.0$ seconds. PD feedback control is turned on again for this returning part of the trajectory. When the returning trajectory is executed, SL main thread running the inverse dynamics computations will continue to keep the arms stable around the resting posture, while another thread is detached to run the ILC update\footnote{Code is available in the public repository \url{https://gitlab.tuebingen.mpg.de/okoc/learning-control} along with the test scripts used to generate the plots in the previous subsections.}. The ILC loop terminates successfully whenever the computed feedforward updates are within the respective torque limits. After a successful termination, if the actual posture is within $0.1$ radians distance of the resting posture, the LQR feedback will be turned on again and the robots will start moving to track the same striking motion. 

We use a simulated ball to make more controlled experiments, focusing on the control aspect in more detail. If the striking robot movements are executed accurately, then the ball in simulation will be returned close to a desired position on the opponent's court. At different points in time we have identified three different sets of link parameters for rigid body dynamics. We can use these parameterizations of rigid body dynamics as potential nominal models to kick-start the learning process. We tested these nominal models first in slowed down hitting movements, where a slow down rate of two means that the number of trajectory points double while the hitting time is held fixed. Cutting down the trajectories to an initial subset of the movement to restrict potential instabilities, or initial masking of some of the joints during ILC updates, are other techniques that we have employed to evaluate these nominal models in a careful manner. Of the three models, only one of them was suitable for the \emph{local} learning that ILC provides. This model is further adapted with the proposed $\alg$ algorithm in order to improve the tracking of the striking movements. Adaptation of the trajectories $\traj_{j}$ and the nominal inputs $\sysInput_{\mathrm{IDM},j}$ was additionally performed on top of ILC, to stabilize the learning process, since an accurate initialization of the joints (especially on the wrist and the elbow) was not possible with the Barrett WAMs. 

We have compared $\alg$ to two other ILC methods: batch ILC \eqref{batch-ilc} and ILC with proportional and derivative (PD) feedback (with constant $p, d$ values). PD type ILC with constant $p$ or $d$ values is often too simplistic, and did not yield any improvement in our setup, even after tuning the $p,d$ values. Batch ILC was tested with ten times downsampled trajectories, with adjustable learning rates. We have found batch ILC to be inferior to the recursive ILC when tested over multiple trajectories (slowed down and cut versions included)\footnote{For batch pseduoinverse-based ILC, inversion of the model matrices \eqref{discreteLTV} around the \emph{unstable} hitting trajectory causes instability, which is alleviated by providing an additional \emph{current iteration ILC} (CI)~\cite{Bristow06}. CI adds the current iteration $k$'s feedback errors to the feedforward compensations for the next iteration $k+1$. As in our preliminary experiments with the Barrett WAM~\cite{Koc15}, we have applied CI in addition to stabilize a downsampled version of batch model-based ILC.}. Recursive ILC without any adaptation is monotonically convergent on average for about five iterations, bringing the root mean squared (RMS) tracking error from about $0.80$ to $0.40$ on average. Repeating the trajectories for five more iterations, we note that the tracking error starts increasing slightly due to introduced oscillations in some of the joints. Introducing adaptation with recursive and cautious ILC (i.e., the proposed approach $\alg$) we can decrease the tracking error further, to about $0.20$ monotonically in five more iterations. This enables a return accuracy of $40 \%$ of the simulated balls to the opponent's court.




The proposed update law $\alg$ evaluated above adapts the discrete-time LTV models with a forgetting factor of $\lambda = 0.8$. This value was chosen experimentally, and could be optimized, e.g., using a dataset of previous ILC performances. The same parameter values are chosen for the initial covariances as in the simulation experiments with the Barrett WAM. Adapting the continuous LTV models, when the trajectories are smoothened suitably with a zero-phase filter, leads to faster updates with similar improvements in tracking performance. Using the online adaptation of the link parameters on the other hand, leads to poorer convergence in tracking for some of the joints (most notably, the elbow). This fact leads us to suspect that the rigid body dynamics model underfits, i.e., the mismatch for our Barrett WAMs is not purely parametric in nature. We see that the final cost (as 2-norm of deviations from desired joint hitting positions and velocities) drops down from $1.70$ to $0.20$ for $\alg$ when the LTV model matrices are adapted directly. After performing ten more iterations, the percentage of balls successfully returned to the opponent's court increases from $40 \%$ to about $60 \%$ on average\footnote{A video showing some example ILC performances for the two robots is available online: \url{https://youtu.be/27vHoLBwLoM}.}. 



\section{Conclusion and Future Work}\label{conclusion}


In this paper we presented a novel Iterative Learning Control (ILC) algorithm that is recursive, cautious and adaptive at the same time. The closed-form update law \eqref{recursive_ilc_cautious} that was presented derives from the adaptive dual control literature and is sometimes referred to as \emph{passive learning}~\cite{Kendrick81}. The algorithm was then recast in a more efficient form (derived in Appendix \ref{derivations}) which does not require the estimation of disturbances and can be implemented as a recursive ILC update. The update law makes it easy to introduce caution with respect to modelling uncertainties and online adaptation of the linearized model matrices. Unlike typical ILC updates, feedback matrices for the tracking of striking trajectories are adapted as well, which are useful for rejecting noise and varying initial conditions. We believe that the introduced ILC update yields a principled approach to adapt the models, as well as their regularizer, based on data.

The proposed algorithm $\alg$ was evaluated in different simulations of increasing complexity. Finally in the last subsection we have presented real robot experiments on our robotic table tennis setup with two Barrett WAMs, see Figure~\ref{ping_pong_moving}. It was shown that the proposed approach leads to an efficient way to learn to track hitting movements online. Hitting movements throughout the experiments are generated in the joint space of the robots and enable them to execute optimal striking motions. Control inputs, as well as a time-varying feedback law, are updated after each trial by using the model-based update rule that considers the deviations from the striking trajectory. After the trajectories are executed, the deviations can be used to adapt the model parameter means and variances using Linear Bayesian Regression (LBR). A forgetting factor was considered in addition to make adaptation more flexible. An adaptation of the reference trajectories as well as the nominal inputs was considered on top of $\alg$ to render the method more effective and stable for initial posture stabilization errors.



Although we have shown a stable and efficient way to learn to track references with ILC, we have not analyzed its generalization to arbitrary trajectories. In our table tennis setup, we are making progress to having the two robots play against each other. Generalization capacity would play an important role in extending the average game duration between the robots, as the trajectories during the table tennis matches would be generated online~\cite{Koc18} according to the state of the game. We believe that in the case where the trajectories are changing, generalizing the learned control commands can be achieved by compressing them to a lower-dimensional input space (i.e., parameters). Learned feedforward commands could be projected to a parameterized feedback matrix, the parameters of which could represent the invariants between the trajectories. An efficient and stable implementation of such parameterizations will be the focus of our future work.



\section*{Acknowledgment}

Part of the research leading to these results has received funding from the European Community's Seventh Framework Programme (FP7-ICT-2013-10) under grant agreement 610878 (3rdHand) and from a project commissioned by the New Energy and Industrial Technology Development Organization (NEDO).

\bibliographystyle{IEEEtran}
\bibliography{./refs}


%
\appendices

\section{Cautious ILC Derivations}\label{derivations}
We provide in this section self-contained derivations of the cautious ILC update rule, given in Equations~\eqref{recursive_ilc_cautious} and simplified in \eqref{recursive_implementation}. Consider the following optimal control problem
\begin{align}
\min_{\linInput} & \, \sum_{j = 1}^{N} \mathbb{E}_{\vec{A}_{j}, \vec{B}_{j}}[\error^{\mathrm{T}}_{k+1,j}\vec{Q}_j\error_{k+1,j} \!+\! \linInput^{\mathrm{T}}_{k+1,j}\vec{R}_j\linInput_{k+1,j}], \\
\textrm{s.t. \ } & \error_{k+1,j+1} = \vec{A}_{j}\error_{k+1,j} + \vec{B}_{j}\sysInput_{k+1,j} + \linDist_{j+1}, \label{dynamics_der}
\end{align}
where the linear time-varying system matrices $\vec{A}_{j}, \vec{B}_{j}$ are random variables with known means and variances. Since $\sysInput_{k+1,j} = \sysInput_{k,j} + \linInput_{k,j}$, we can rewrite \eqref{dynamics_der} as
\begin{align}
 &\error_{k+1,j+1} = \vec{A}_{j}\error_{k+1,j} + \vec{B}_{j}\linInput_{k,j} + \bar{\linDist}_{j+1}, \label{pred_error}\\
&\bar{\linDist}_{j+1} = \vec{B}_{j}\sysInput_{k,j} + \linDist_{j+1}.
\end{align}
\noindent The iteration index $k$ will be removed until the last subsection due to space constraints. Notice that the Value Function for the optimal control problem \eqref{dynamics_der} is a quadratic function of the errors along the trajectory, 
\begin{align}
\ValueFunction(\error,j) = \error^{\mathrm{T}}\ricMat_{j}\error + 2\error^{\mathrm{T}}\ricVecOne_j + \ricScalar_j, \label{quadraticValueFnc}
\end{align}
\noindent for time-varying matrices $\ricMat_{j} \in \mathbb{R}^{2n \times 2n}$, vectors $\ricVecOne_j \in \mathbb{R}^{2n}$ and $\ricScalar_j \in \mathbb{R}$. We can then apply dynamic programming to compute the optimal solution recursively
\small
\begin{equation}
\begin{aligned}
&\ValueFunction(\error_j,j) = \min_{\linInput_j} \big(\error_j^{\mathrm{T}}\vec{Q}_j\error_j + \linInput_j^{\mathrm{T}}\vec{R}_j\linInput_j + \ValueFunction(\error_{j+1},j+1)\big) \label{value_fnc_der}, \\
&\ValueFunction(\error_{j+1},j+1) = \mathbb{E}_{\vec{A}_j,\vec{B}_j}[ 2\ricVecOne_{j+1}^{\mathrm{T}}(\vec{A}_j\error_j + \vec{B}_j\linInput_j + \bar{\linDist}_{j+1}) +\ricScalar_{j+1} \\
& \phantom{\hspace{5mm}} + (\vec{A}_j\error_j + \vec{B}_j\linInput_j + \bar{\linDist}_{j+1})^{\mathrm{T}}\ricMat_{j+1}(\vec{A}_j\error_j + \vec{B}_j\linInput_j + \bar{\linDist}_{j+1})].
\end{aligned}
\end{equation}
\normalsize
\noindent The recursion starts from $\ricMat_{N} = \vec{Q}_N$. Taking derivative w.r.t. $\linInput_j$ of the right-hand side, we get
\begin{equation}
\begin{aligned}
&\vec{R}_j\linInput_j + \big(\mathbb{E}_{\vec{A}_j,\vec{B}_j}[\vec{B}_j^{\mathrm{T}}\ricMat_{j+1}\vec{A}_j]\error_j + \mathbb{E}_{\vec{B}_j}[\vec{B}_j^{\mathrm{T}}\ricMat_{j+1}\vec{B}_j]\linInput_j + \\ &\mathbb{E}_{\vec{B}_{j}}[\vec{B}_{j}^{\mathrm{T}}(\ricMat_{j+1}\bar{\linDist}_{j+1} + \ricVecOne_{j+1})]\big)  = 0.
\end{aligned}
\label{optimal_control_app}
\end{equation}
Solving \eqref{optimal_control_app} for the optimal control input compensations, and arranging using the notation in \eqref{recursive_ilc_cautious}
\begin{equation}
\begin{aligned}
\linInput_j &= \fbMat_j\error_j - \matOne_j^{-1}\ffVec_j, \\
\fbMat_j &= -\matOne_j^{-1}\matTwo_j, \\
\matOne_j &= \vec{R}_j + \mathbb{E}_{\vec{B}_j}[\vec{B}_j^{\mathrm{T}}\ricMat_{j+1}\vec{B}_j], \\
\matTwo_j &= \mathbb{E}_{\vec{A}_j,\vec{B}_j}[\vec{B}_j^{\mathrm{T}}\ricMat_{j+1}\vec{A}_j], \\
\ffVec_j &= \mathbb{E}_{\vec{B}_j}[\vec{B}_j^{\mathrm{T}}(\ricMat_{j+1}\bar{\linDist}_{j+1} + \ricVecOne_{j+1})]. 
\end{aligned}
\label{opt_control_der}
\phantom{\hspace{6cm}} 
\end{equation}
In order to derive a Riccati-like equation, we plug \eqref{opt_control_der} into \eqref{value_fnc_der}, and using \eqref{quadraticValueFnc} get
\begin{equation}
\begin{aligned}
&\error^{\mathrm{T}}\ricMat_{j}\error \!+\! 2\error^{\mathrm{T}}\ricVecOne_j \!+\! \ricScalar_j = \error_j^{\mathrm{T}}\vec{Q}_j\error_j \!+\! \error_j^{\mathrm{T}}(\matTwo^{\mathrm{T}}_j\matOne_j^{-1}\vec{R}_j\matOne_j^{-1}\matTwo_j)\error_j \\ 
& \phantom{\hspace{2cm}} + 2\ffVec_j^{\mathrm{T}}\matOne_j^{-1}\vec{R}_j\matOne_j^{-1}\matTwo_j\error_j + \ffVec_j^{\mathrm{T}}\matOne_j^{-1}\vec{R}_j\matOne_j^{-1}\ffVec_j \\
& \phantom{\hspace{2cm}} + \mathbb{E}_{\vec{A}_j,\vec{B}_j}[(\bar{\vec{A}}_j\error_j + \vec{m}_{j})^{\mathrm{T}}\ricMat_{j+1}(\bar{\vec{A}}_j\error_j + \vec{m}_{j})] \\
& \phantom{\hspace{2cm}} + 2\mathbb{E}_{\vec{A}_j,\vec{B}_j}[(\bar{\vec{A}}_j\error_j + \vec{m}_{j})^{\mathrm{T}}\ricVecOne_{j+1}] + \ricScalar_{j+1}, 
\end{aligned}
\phantom{\hspace{6cm}}
\end{equation}
\noindent where we have introduced the terms
\begin{equation}
\begin{aligned}
\bar{\vec{A}}_j &= \vec{A}_j + \vec{B}_j\fbMat_j, \\
\vec{m}_{j} &= \bar{\linDist}_{j+1} - \vec{B}_{j}\matOne_{j}^{-1}\ffVec_{j}.
\end{aligned}
\phantom{\hspace{6cm}}
\end{equation}
\noindent Checking for the equality of the quadratic terms we get, after some cancellations,
\begin{equation}
\begin{aligned}
\ricMat_{j} &= \vec{Q}_j + \matThree_j - \matTwo^{\mathrm{T}}_j\matOne_j^{-1}\matTwo_j, \\
\matThree_j &= \mathbb{E}_{\vec{A}_j}[\vec{A}_j^{\mathrm{T}}\ricMat_{j+1}\vec{A}_j], \\
\ricVecOne_{j} &= \matTwo_j^{\mathrm{T}}\matOne_j^{-1}\vec{R}_j\matOne_j^{-1}\ffVec_{j} + \mathbb{E}_{\vec{A}_j,\vec{B}_j}[\bar{\vec{A}}_j^{\mathrm{T}}(\ricMat_{j+1}\vec{m}_{j} + \ricVecOne_{j+1})].
\end{aligned}	
\label{riccati_der}
\phantom{\hspace{6cm}}
\end{equation}
\subsubsection{Rewriting the feedforward recursion} The control input compensations calculated in \eqref{opt_control_der} can be simplified significantly by noting that the last three terms in the feedforward recursion of \eqref{riccati_der} 
\begin{equation}
\begin{aligned}
\ricVecOne_{j} &= \mathbb{E}[\bar{\vec{A}}_j^{\mathrm{T}}(\ricVecOne_{j+1} \!+\! \ricMat_{j+1}\bar{\linDist}_{j+1})] \!-\! \mathbb{E}[\vec{A}_{j}^{\mathrm{T}}\ricMat_{j+1}\vec{B}_j]\matOne_{j}^{-1}\ffVec_{j} \\ &-\fbMat_{j}^{\mathrm{T}}\mathbb{E}[\vec{B}_{j}^{\mathrm{T}}\ricMat_{j+1}\vec{B}_j]\matOne_{j}^{-1}\ffVec_{j} - \fbMat_{j}^{\mathrm{T}}\vec{R}_j\matOne_j^{-1}\ffVec_{j},
\end{aligned}
\end{equation}
\noindent cancel out, leaving
\begin{equation}
\begin{aligned}
\ricVecOne_{j} &= \mathbb{E}_{\vec{A}_j,\vec{B}_j}[\bar{\vec{A}}_j^{\mathrm{T}}(\ricVecOne_{j+1} \!+\! \ricMat_{j+1}\bar{\linDist}_{j+1})].
\end{aligned}
\label{riccati_ff_der}
\end{equation}
The cancellations can be seen easily by rewriting the first term in terms of the feedback matrix and grouping the last two terms together
\begin{equation}
-\fbMat^{\mathrm{T}}_{j}\matOne_{j}\matOne_{j}^{-1}\ffVec_{j} + \fbMat^{\mathrm{T}}_{j}\ffVec_{j} = 0.
\end{equation}
\subsubsection{Simplifying the feedforward recursion} The feedforward recursion in \eqref{riccati_ff_der} still requires the explicit estimation of disturbances. This equation can be simplified further by rewriting the disturbances in terms of the previous trial errors
\begin{equation}
\begin{aligned}
\bar{\linDist}_{j+1} &= \error_{k,j+1} - \vec{A}_{j}\error_{k,j}, \\
\ffVec_j &= \mathbb{E}[\vec{B}_j^{\mathrm{T}}\big(\ricMat_{j+1}\error_{k,j+1} + \ricVecOne_{j+1}\big)] - \matTwo_{j}\error_{k,j}. 
\end{aligned}
\end{equation}
Introducing $\ricVecTwo_{j+1} = \ricMat_{j+1}\error_{k,j+1} + \ricVecOne_{j+1}$, we can rewrite the optimal control input compensations as
\begin{equation}
\begin{aligned}
\linInput_j &= \fbMat_j(\error_{k+1,j} - \error_{k,j}) - \matOne_j^{-1}\mathbb{E}[\vec{B}_{j}^{\mathrm{T}}\ricVecTwo_{j+1}].
\end{aligned}
\label{riccati_ff_der2}
\end{equation}
Rewriting \eqref{riccati_ff_der} in terms of $\ricVecTwo_{j}$, we get
\begin{equation}
\begin{aligned}
\ricVecTwo_{j} = \mathbb{E}[\bar{\vec{A}}_{j}^{\mathrm{T}}\ricVecTwo_{j+1}] + \left(\ricMat_{j} - \mathbb{E}[(\vec{A}_{j} + \vec{B}_{j}\fbMat_{j})^{\mathrm{T}}\ricMat_{j+1}\vec{A}_{j}]\right)\error_{k,j},
\end{aligned}
\end{equation}
since $\ricMat_{j} = \vec{Q}_j + \matThree_j - \matTwo^{\mathrm{T}}_j\matOne_j^{-1}\matTwo_j$, the last term becomes
\begin{equation}
\begin{aligned}
\left(\ricMat_{j} - \matThree_{j} - \fbMat_{j}^{\mathrm{T}}\matTwo_{j}\right)\error_{k,j} = \vec{Q}_{j}\error_{k,j}, 
\end{aligned}
\end{equation}
hence, the feedforward recursion defining \eqref{riccati_ff_der2} can be computed independently of disturbance estimates
\begin{equation}
\begin{aligned}
\ricVecTwo_{j} = \mathbb{E}[\bar{\vec{A}}_{j}^{\mathrm{T}}\ricVecTwo_{j+1}] + \vec{Q}_{j}\error_{k,j}, \ j = 1, \ldots, N-1, 
\end{aligned}
\end{equation}
\noindent starting from $\ricVecTwo_{N} = \vec{0}$.

\section{Movement Generation for Table Tennis}\label{hitting}
In a highly dynamic and complex task such as robot table tennis, one often needs to consider an extension of the standard trajectory tracking task. Based on the varying initial positions and velocities of the robot arm and the trajectory of the incoming ball, in each table tennis stroke the robot arm needs to track different trajectories that start from different initial conditions and end with different desired goal states of the arm. Moreover these trajectories need to be optimized in time to intercept the ball. The striking trajectories $\traj(t) = [\joint_{\text{des}}(t),\dot{\joint}_{\text{des}}(t)]^{\mathrm{T}}$ are generated online and tracked using the proposed ILC approach.

Striking movement primitives suited to table tennis have been proposed in \cite{Kober10} and \cite{Muelling13} as an extension of discrete Dynamic Movement Primitives (DMP). Unlike the original formulation~\cite{Ijspeert02}, these extensions allow for an arbitrary velocity profile to be attached to the primitives around hitting time. However, these approaches are heavily structured for the problem at hand, introducing and tuning additional domain parameters. In~\cite{Koc15} we instead proposed to use rhythmic movement primitives that allow for a limit cycle attractor, which is desirable if we want to maintain the striking motion through goal state. After the striking is completed the DMP can be used to return back to initial state or it can be terminated by setting the forcing terms to zero. An example is shown in Figure~\ref{tracking-with-pd}.

One of the problems with such (kinesthetic) teach-in based approaches is that it is difficult
to train heavy robots well for successful performance. For example, the shoulder of the Barrett WAM arm shown in Figure~\ref{robot} weighs 10 kg alone. It is rather difficult for humans to move the links with heavy inertia. The slower movements of the heavier links are typically compensated with faster movements of the lighter links (such as the wrist). However, tracking these trajectories can also be harder for more demanding wrist movements. An additional difficulty with cable-driven robots such as the Barrett WAM is that the wrists are harder to control. 

Based on these considerations, we have worked on a free-final time optimal control based approach to generate minimum acceleration hitting movements for table tennis~\cite{Koc18}. In the experiments section, we focus on learning to track these hitting movements. These trajectories are third order polynomials for each degree of freedom of the robot. 

We will briefly introduce here the trajectory generation framework introduced in~\cite{Koc18}. Consider the following \emph{free-time} optimal control problem~\cite{Liberzon11}
\begin{align}
\min_{\ddot{\joint},T} & \int\limits_{0}^{T} \ddot{\joint}(t)^{\mathrm{T}}\vec{R}\ddot{\joint}(t) \ \mathrm{d}t \label{costFnc1} \\
\text{s.t. }  
&\hitFun\big(\joint(\hitTime),\hitTime\big) \in \hit, \label{hitFunc} \\
&\netFun\big(\joint(\hitTime),\dot{\joint}(\hitTime),\hitTime\big) \in \net, \label{netFunc} \\
&\landFun\big(\joint(\hitTime),\dot{\joint}(\hitTime),\hitTime\big) \in \landEvent, \label{landFunc} \\
& \joint(0) = \joint_{0}, \label{initCond1} \\
& \dot{\joint}(0) = \dot{\joint}_{0}, \label{initCond2}
\end{align}
\noindent where the final hitting time $\hitTime$ is an additional variable to be optimized along with the joint accelerations $\ddot{\joint}(t) \colon [0,\hitTime] \to \mathbb{R}^{n}$. The weighting matrix $\vec{R}$ for the accelerations is positive definite. 
Initial conditions for the robot are the joint positions $\joint_0$ and joint velocities $\dot{\joint}_0$.
The inequality constraints \mbox{\eqref{hitFunc} -- \eqref{landFunc}} ensure that the task requirements for table tennis are satisfied, namely, hitting the ball, passing the net, and landing on the opponent's court. 

Solutions of \mbox{\eqref{costFnc1} -- \eqref{initCond2}} can be found using Pontryagin's minimum principle~\cite{Pontryagin}. The optimal $\joint(t)$ in both cases is a third degree polynomial for each degree of freedom. The striking time $T$, the joint position and velocity values at striking time $\joint_f$ and $\dot{\joint}_f$ fully parametrize this problem. The time it takes to return to the starting posture, $T_{\mathrm{rest}}$ can be chosen suitably, e.g., based on the speed of the ballgun. The polynomial coefficients for the striking trajectory
\begin{align}
\joint_{\mathrm{strike}}(t) &= \vec{a}_3 t^3  + \vec{a}_2 t^2 + \dot{\joint}_0 t + \joint_0, \label{strike_traj}
\end{align} 
\noindent can then be fully determined in joint-space
\begin{gather}
\begin{aligned}
\vec{a}_3 &= \frac{2}{T^3}(\joint_0 - \joint_f) + \frac{1}{T^2}(\dot{\joint}_0 + \dot{\joint}_f), \\
\vec{a}_2 &= \frac{3}{T^2}(\joint_f - \joint_0) - \frac{1}{T}(\dot{\joint}_f + 2\dot{\joint}_0),
\label{coeffs}
\end{aligned}
\end{gather}
\noindent for each degree of freedom of the robot.

%

%
%
%




\end{document}